# Transformers in Healthcare: A Survey

Subhash Nerella[1]*, Sabyasachi Bandyopadhyay[1]*, Jiaqing Zhang[2], Miguel Contreras[1], Scott Siegel[1], Aysegul Bumin[3], Brandon Silva[3], Jessica Sena[4], Benjamin Shickel[5], Azra Bihorac[5], Kia Khezeli[1], Parisa Rashidi[1].

[1]J. Crayton Pruitt Family Department of Biomedical Engineering, University of Florida, Florida, USA.

[2]Department of Electrical and Computer Engineering, University of Florida, Florida, USA.

[3]Department of Computer and Information Science and Engineering, University of Florida, Florida, USA.

[4]Department of Computer Science, Universidad Federal de Minas Gerais, Belo Horizonte, Brazil

[5]Department of Medicine, University of Florida, Gainesville, FL, United States

*These authors contributed equally to the manuscript.

## Abstract

With Artificial Intelligence (AI) increasingly permeating various aspects of society, including healthcare, the adoption of the Transformers neural network architecture is rapidly changing many applications. Transformer is a type of deep learning architecture initially developed to solve general-purpose Natural Language Processing (NLP) tasks and has subsequently been adapted in many fields, including healthcare. In this survey paper, we provide an overview of how this architecture has been adopted to analyze various forms of data, including medical imaging, structured and unstructured Electronic Health Records (EHR), social media, physiological signals, and biomolecular sequences. Those models could help in clinical diagnosis, report generation, data reconstruction, and drug/protein synthesis. We identified relevant studies using the Preferred Reporting Items for Systematic Reviews and Meta-Analyses (PRISMA) guidelines. We also discuss the benefits and limitations of using transformers in healthcare and examine issues such as computational cost, model interpretability, fairness, alignment with human values, ethical implications, and environmental impact.

## 1 Introduction

The last decade has seen an explosion in data generation in healthcare practices. Healthcare data accounts for 30% of the global data ecosystem and is expected to grow in the coming years [1]. Due to this trend, the last decade has witnessed a simultaneous burgeoning of machine learning/deep learning algorithms used for combing through large healthcare datasets to facilitate diagnosis, prognosis, and decision-making.

Transformer [2] is a type of Deep Neural Network (DNN) introduced in 2017 for sequence modeling problems, especially in the Natural Language Processing (NLP) domain [3]. Before the introduction of the Transformer [2], the most popular deep learning architectures, such as recurrent neural networks (RNNs) [4], and their variants worked in a sequential fashion which precluded parallelization during training, substantially increasing the training time. In contrast, transformers employ a "Scaled Dot-Product Attention" mechanism that is parallelizable. This unique attention mechanism allows for large-scale pretraining. Additionally, self-supervised pretraining paradigm such as masked language modeling onlarge unlabeled datasets enabled transformers to be trained without costly annotations.

Transformer model, although originally designed for the NLP [3] domain, Transformers have witnessed adaptations in various domains such as computer vision [5, 6], remote sensing [7], time series [8], speech processing [9] and multimodal learning [10]. Consequently, modality specific surveys emerged, focusing on medical imaging [11-13] and biomedical language models [14] in the medical domain. This paper aims to provide comprehensive overview of Transformer models utilized across multiple modalities of data to address healthcare objectives. We discuss pre-training strategies to manage the lack of robust and annotated healthcare datasets. The rest of the paper is organized as follows: Section 2 discusses the strategy to search for relevant citations; Section 3 describes the architecture of the original transformer; Section 4 describes the two primary Transformer variants: the Bidirectional Encoder Representations from Transformers (BERT) and the Vision Transformer (ViT). Section 5 describes advancements in large language models (LLM), and section 6 through 12 provides a review of Transformers in healthcare. Finally, section 13 discusses limitations, interpretability, environmental impact, computational costs, bias, and fairness.

## 2 Search Strategy and Selection criteria

We used Google Scholar and PubMed search engines to search for Transformer studies in healthcare. Since Vaswani et al.'s initial Transformer network was published in 2017, we limited our search to studies published after 2017. The search was divided into six categories: **clinical NLP, EHR, social media, medical imaging, biomolecules, and bio-physical signals**. We utilized PRISMA guidelines shown in Fig 1 to find relevant studies and report our findings.

For each category, we used the terms "health" or "medical" or "clinical" to focus the search on the healthcare domain. Finally, each category used a precise set of keywords unique to that domain. The keywords are combined with logical operators such as "AND" and "OR" to enhance the search results quality. A detailed list of search queries can be found in Table 1. We used Harzing's Publish or Perish [15] to retrieve studies and Covidence [16] to perform PRISMA analysis on the retrieved studies.

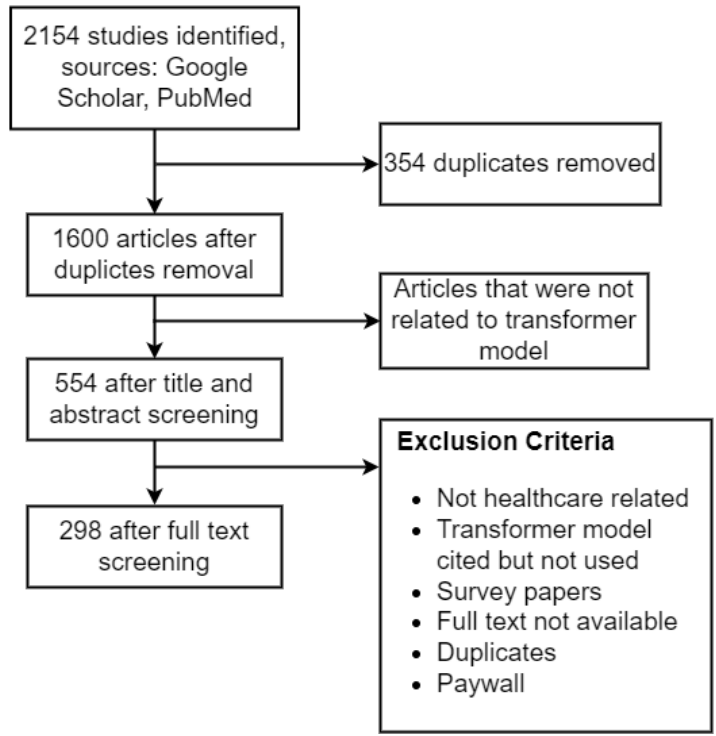

Figure 1. Flow diagram depicting the PRISMA analysis process for selecting relevant studies for inclusion and exclusion.

Table 1. Search queries used to extract relevant studies for each topic

| Topic | Search query |
|---|---|
| Clinical NLP | ("coreference" OR ("semantic textual similarity" OR STS) OR ("named entity recognition" OR NER) OR "relation extraction" OR "natural language inference" OR "question answering" OR "entity normalization") AND (BERT OR Transformer) AND ("clinical" OR "medical" OR "biomedical" OR "EHR") from 2017 |
| Medical Imaging | (Segmentation OR registration OR "image captioning" OR "report generation" OR "visual question answering" OR "image synthesis" OR "classification" OR "reconstruction") AND ("Transformer" OR "vision transformer") AND ("clinical" OR "medical" OR "biomedical" OR "EHR") from 2017 |
| Critical Care | (Transformer) AND ("deep learning" OR "machine learning") AND ("critical care" OR "surgery" OR "surgical") from 2017 |
| Structured EHR | (Transformer OR BERT) AND ("deep learning" OR "machine learning") AND (EHR OR "electronic health records") from 2017 |
| Social Media | (Transformer OR BERT) AND ("deep learning" OR "machine learning") AND ("social media" OR "crowdsource" OR "crowdsourcing" OR "twitter" OR "tweet") from 2017 |
| Bio-physical Signals | (Transformer OR BERT) AND ("deep learning" OR "machine learning") AND ("medical" OR "health" OR "clinical" OR "biomedical") AND ("signal" OR "ECG" OR "EMG" OR "EEG" OR "human activity" OR "HAR") from 2017 |
| Biomolecular Sequences | (Transformer OR BERT) AND ("deep learning" OR "machine learning") AND (DNA OR RNA OR gene OR genome OR genomic OR transcriptomic OR protein OR proteomic OR metabolite OR metabolism OR metabolomic OR chromosome OR receptor OR mitochondria OR splicing) from 2017 |

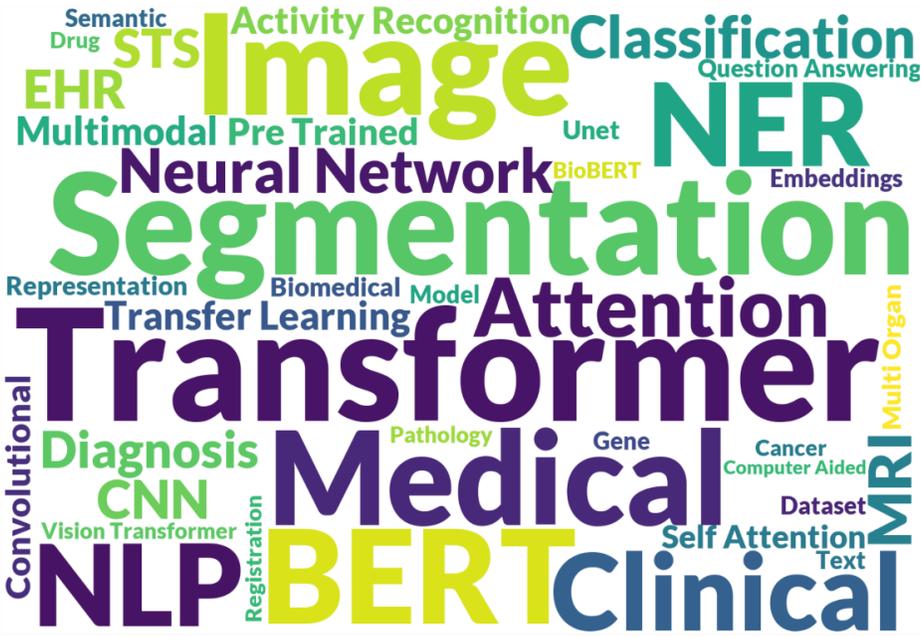

Figure 2. Word cloud depiction of keywords used in the surveyed literature. Abbreviations. BERT; Bidirectional Encoder Representations from Transformers, CNN; Convolutional Neural Networks, EHR; Electronic Health Records, MRI; Magnetic Resonance Imaging, NER; Named Entity Recognition, NLP; Natural Language Processing, STS; Semantic Textural Similarity

We identified the top keywords to provide an overview of key concepts, data modalities, and tasks. The word cloud in Fig. 2 shows the 50 most common keywords across articles, with a larger font representing more papers; while Fig. 3 shows data modalities and the corresponding tasks.

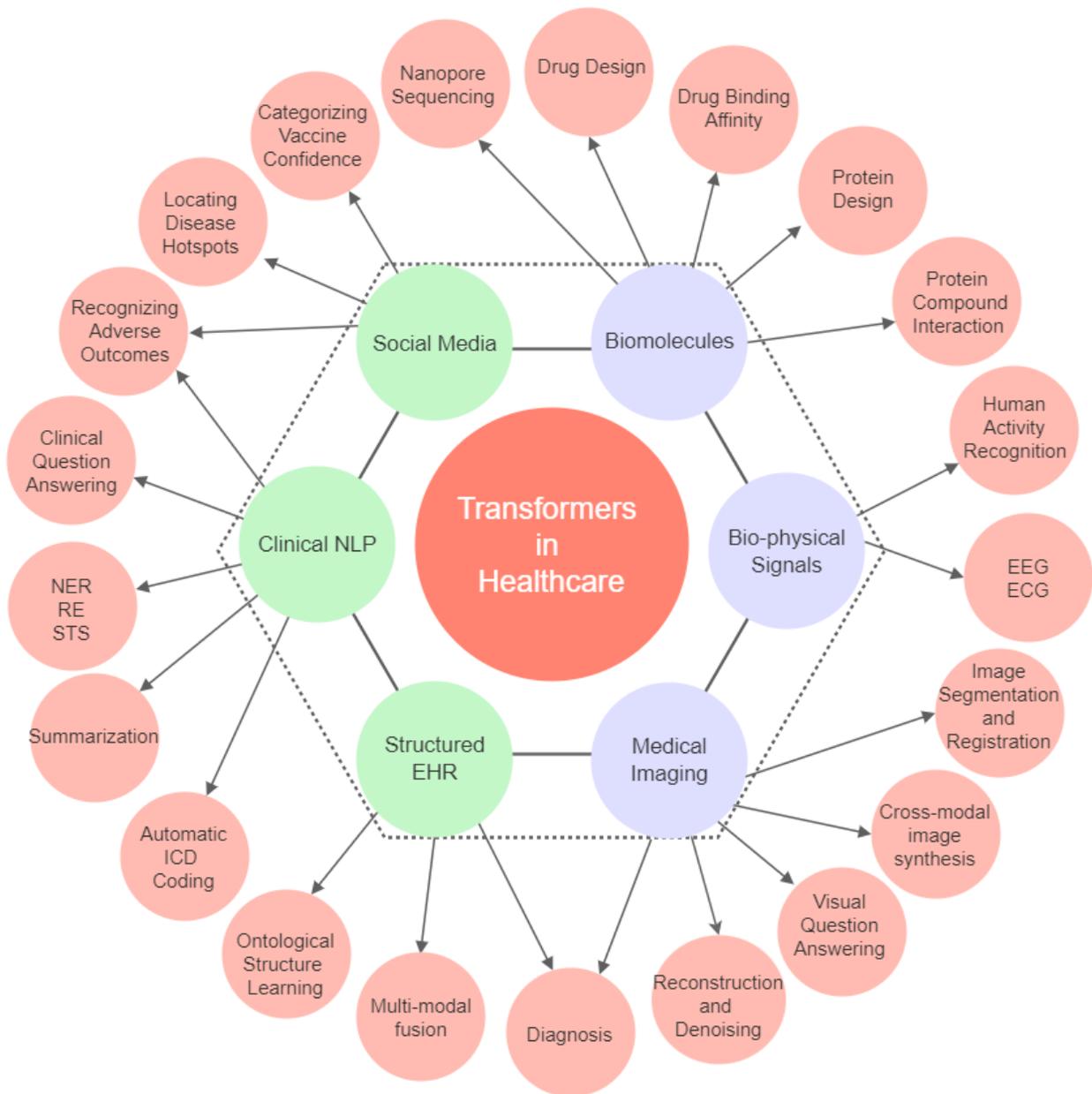

Figure 3. Major healthcare data source modalities and corresponding tasks. Abbreviations: EEG; Electroencephalography, ECG; Electrocardiogram, NER; Named Entity Recognition, RE; Relation Extraction, STS; Semantic Textual Similarity.

## 3   Background

Transformers are multilayered neural networks formed by stacking multiple encoder-decoder blocks that utilize the attention mechanism, as explained in the following section.

### 3.1   Attention

The attention mechanism computes the similarity between individual input tokens, such as the vectors of word embeddings. In a basic Transformer architecture, each input embedding generally can take three roles: (1) Query $Q$ as the current focus of attention when being compared to all of the other input tokens, (2) Key $K$ as a input token being compared to the current focus of attention, and (3) Value $V$ as a value used to compute the output for the current focus of attention. The attention function can be considered a mapping between a query and a set of key-value pairs to produce an output [2].

We will represent the input $X \in R^{n \times d}$ as a sequence of $n$ tokens with an embedding dimension of $d$. The input sequence $X$ is linearly transformed into query $Q$, key $K$, and value $V$ using equations 1, 2, and 3, respectively.

$$Q = X \cdot W_q \tag{1}$$

$$K = X \cdot W_k \tag{2}$$

$$V = X \cdot W_v \tag{3}$$

where $W_q$, $W_k$, and $W_v$ are the weight matrices to obtain query, key, and value matrices. The query, key, and value are then used in Equation 4, representing the scaled dot product attention operation (layer $R^{n \times d_v}$ in Fig. 4b).

$$Attention(Q, K, V) = softmax(\alpha Q \cdot K^T) \cdot V \tag{4}$$

In Equation 4, a scaled dot product operation is performed between the query and key matrices, followed by a Softmax function. The scale factor $\alpha$ is used to mitigate the vanishing gradient problem and numerical instability and is typically chosen to be $1 / (\sqrt{d_k})$ where $d_k$ is the key dimension.

### 3.2 Attention Mechanisms

Transformer models primarily use three types of attention: self-attention, masked self-attention, and cross-attention.

#### 3.2.1 Self-Attention

Self-attention is when attention is computed between tokens in the same sequence. The self-attention block is found in the Transformer encoder. The dimensions of query, key, and value are the same in self-attention, i.e., $d_k = d_q = d_v$.

#### 3.2.2 Masked Self-Attention

In sequence prediction problems, such as machine translation, the context of previous tokens $i = 0 \dots j$ in a sequence is used to predict the subsequent output. The desired output can then be provided as an input to the Transformer architecture to achieve sequence-to-sequence decoding. A mask is typically employed to prevent the model from attending to subsequent tokens in a sequence. The mask $M$ (Equation 5) is a square upper triangular matrix with dimension $n$, where $n$ is the number of tokens in the input sequence.

$$M_{ij} = -\infty \text{ if } i < j \text{ else } 0 \tag{5}$$

The mask is applied to the scaled dot product of the query and key via element-wise addition, as in Equation 6.

$$\text{Masked Attention}(Q, K, V) = softmax\left(\frac{Q \cdot K^T}{\sqrt{d_k}} + M\right) V \qquad (6)$$

### 3.2.3 Cross-Attention

Cross-attention is attention computed between tokens of one sequence with tokens of another sequence. In Transformer, the input and desired output sequences interact through cross-attention in the decoder module. The cross-attention module receives queries from the previous masked self-attention layer of the decoder and the keys and values from the last encoder. Queries correspond to the desired output sequence, while the keys and values are generated based on the input sequence in the encoder.

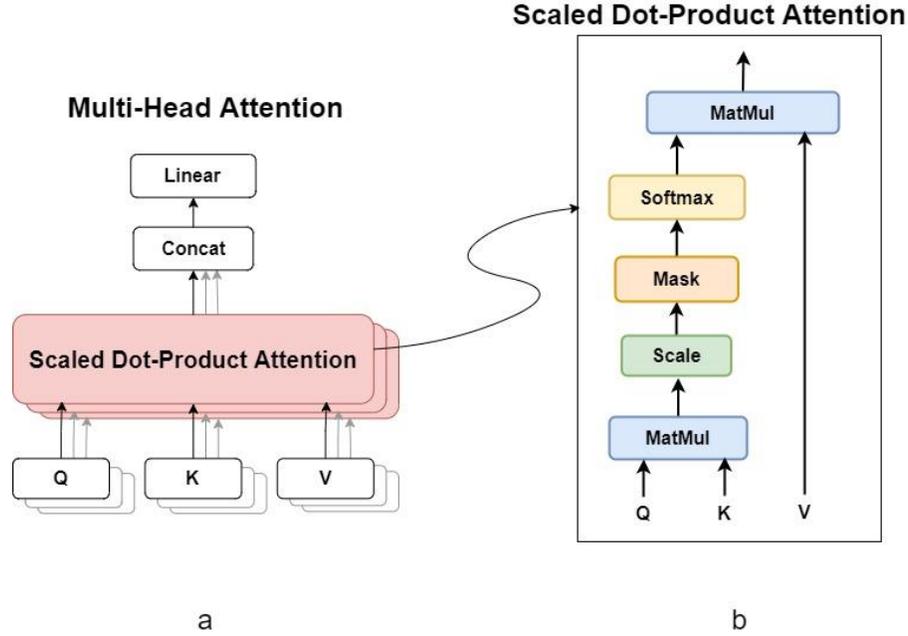

Figure 4. Multi head attention mechanism. In the encoder and decoder, multiple attention heads are stacked together and their outputs are concatenated

### 3.2.4 Multi-Head Attention

It has been shown that multiple attention operations compared to a single attention computation, can improve the model's performance by capturing different similarity relationships in the sequence [2]. The attention blocks in both the encoder and decoder are computed with $h$ attention heads, as shown in Fig 4. The original Transformer model employed $h$ = 8 attention heads. Every attention head has three learnable weight matrices. $W_q^i$, $W_k^i$, and $W_v^i$ where $i$ represents a particular attention head. The attention outputs from multiple heads are then concatenated and linearly transformed to the model dimension with a parameter matrix $W_o$. Multi-head self-attention block can be represented by the equations 7 and 8.

$$head_i = Attention(X \cdot W_q^i, X \cdot W_k^i, X \cdot W_v^i) \qquad (7)$$

$$MHSA = Concat(head_0, head_1, \ldots, head_{h-1}) \cdot W_o \qquad (8)$$

## 3.3 Position-wise Feed-Forward Network

The output of the attention modules is passed to a two-layered feedforward network (FFN). The FFN performs an independent position-wise operation on each entity of the sequence. Parameters of this network are shared across all positions of the sequence.

Let $\mathcal{H}$ be the output of the multi-head attention block and $d_m$ be the model dimension. The first linear layer transforms $\mathcal{H}$ from *dimension* $d_m$ to an intermediate dimension $d_f$, also referred to as the feedforward dimension. The second linear layer transforms the output of the first linear layer from $d_f$ to the original model dimension $d_m$. The FFN is given by equation 9.

$$\mathcal{F}(\mathcal{H}) = ReLU(\mathcal{H} \cdot W_1 + b_1) \cdot W_2 + b_2 \tag{9}$$

The intermediate dimension $d_f$, is usually set to a value larger than $d_m$.

## 3.4 Residual Connections and Layer Normalization

Residual connections [17] allow gradients to skip non-linear activation functions, followed by Layer Normalization [18]. Layer Normalization scales the values of all hidden layers to a similar range to avoid exploding or diminishing values obtained through a chain of multiplication operations.

## 3.5 Positional Encodings

Because the self-attention module attends to all tokens of a sequence in parallel, it intrinsically neglects the order of tokens in the sequence. This necessitates using a positional encoding (PE) vector that denotes the unique position of each token. Transformers use a combination of sine and cosine functions of different frequencies to create PE vectors shown in equation 10. PE vectors are added to the embeddings of each input token; therefore PE dimension is chosen to be the same as the embedding dimension. Since sine and cosine functions have values in the range [-1, 1], the values of the positional encoding matrix are constrained to a normalized range. This technique enables Transformers to capture the relationship between items that are both close and far from one another in a sequence.

$$PE_{(pos,i)} = \begin{cases} \sin(pos \cdot \omega_k), & if\ i = 2k \\ \cos(pos \cdot \omega_k), & if\ i = 2k+1 \end{cases} \tag{10}$$

$$\omega_k = \frac{1}{10000^{2k/d}}, \quad k = 1, 2, \ldots, \frac{d}{2}$$

In equation 10, $d$ is the PE dimension, $i$ is the index along PE dimension, and $pos$ is the element's position in the input sequence. PE is added to the token embeddings based on the position therefore PE dimension is chosen to be same as the embedding dimension.

## 3.6 Assembling a Transformer

Transformer consists of an encoder and a decoder network. The encoder consists of identical encoder blocks stacked upon each other, each consisting of a self-attention and an FFN layer. The decoder consists of stacked identical decoder blocks, each consisting of a masked self-attention layer, cross-attention layer, and FFN layer. The encoder transforms an input sequence into encoded representations, while the decoder operates upon these representations.

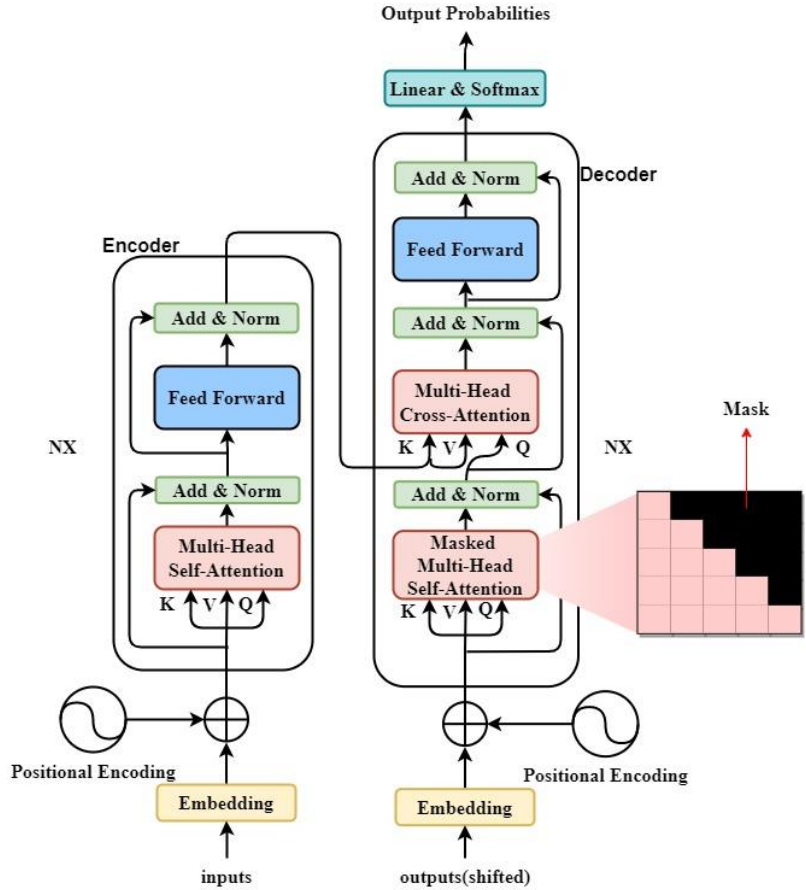

Fig 5. Schematic of the Transformer architecture [2, 19].

Encoder block in a transformer can be expressed as:

$$X_{int} = LN(MHSA(X)) + X \tag{11}$$

$$Z = LN(FFN(X_{int})) + X_{int} \tag{12}$$

Decoder block in a transformer can be expressed as:

$$Y_{int} = LN(MHMSA(Y)) + Y \tag{13}$$

$$Y_{int} = LN(MHCA(Y_{int}, Z)) + Y_{int} \tag{14}$$

$$Out = LN(FFN(Y_{int})) + Y_{int} \tag{15}$$

MHSA: Multi-head self-attention, MHMSA: Multi-head masked self-attention, LN: Layer norm, FFN: Feed forward network, MHCA: Multi-head cross attention. Equations 11-15 have layer norm ($LN$) and residual connections. $X$ and $Y$ represent input and desired output sequence respectively. $X_{int}$ and $Y_{int}$ represent intermediate outputs within encoder and decoder blocks respectively.

The original Transformer architecture (Vaswani et al., 2017), shown in Fig 5, had six identical stacked encoders and six identical stacked decoder blocks. Each encoder block comprised multi-head self-attention followed by FFN. Every decoder block consists of multi-head masked self-attention, multi-head cross-attention, and FFN arranged sequentially. The cross-attention layers attend to queries from the previous masked attention layers, whereas keys and values are obtained from the output of the final encoder block. The output of the last encoder is used to obtain the keys and values to compute the multi-head cross attention in all the decoder layers.

### 3.7 Computational Complexity of Transformer Attention

Unlike traditional neural networks, which require fixed input sizes, the self-attention mechanism can attend to variable-length input sizes. The Transformer attention has $O(n^2 \cdot d)$ time complexity where $n$ and $d$ are the input sequence length and the model dimension. For long input sequences, this attention computation becomes computationally expensive. Many Transformer variants try to reduce the computational complexity via different approaches [20].

### 3.8 Transformer Model Usage

In general, Transformer architectures can be divided into three categories.

- Encoder-Decoder: consists of multiple encoders and decoders blocks and is typically used in sequence-to-sequence modeling tasks, such as machine translation.
- Encoder only: Only the encoder blocks are used to model the input sequence. The output of the encoder is a contextual representation of the input sequence. This type of architecture is used for classification or label prediction problems (most models in this review).
- Decoder only: Only decoder blocks are used. This architecture is used for sequence generation, image captioning, and language modeling tasks.

## 4 Mainstream Transformer-based Architectures

In this section, we will discuss the two prominent transformer-based architectures with significant impact on NLP and computer vision.

### 4.1 Bidirectional Encoder Representations from Transformers (BERT)

BERT [21], Fig 6, is an encoder-only Transformer architecture that can produce rich contextualized word/sentence embeddings for NLP. Unlike traditional language models, which read text input sequentially (left-to-right or right-to-left), the Transformer encoder in BERT reads the entire sequence of words at once, thereby learning a richer representation of context and information flow in a sentence. The BERT architecture uses self-supervised pretraining steps, namely Masked Language Modeling (MLM), to create context-sensitive word embeddings, and Next Sentence Prediction (NSP) to model sequential association between sentences. MLM masks a fraction of the input tokens and aims to predict them based on their context. This helps to disentangle ambiguity in the text by using surrounding text to establish context. In NSP, a combination of two sentences is fed to the Transformer encoder. In 50% of cases, the second sentence is the next sentence in the original text, while in the remaining 50% of cases, the second sentence is randomly selected. The encoder learns to distinguish scenarios where the sentences are logically linked. When training the BERT model, MLM and NSP are trained together to minimize the combined loss function of the two strategies.

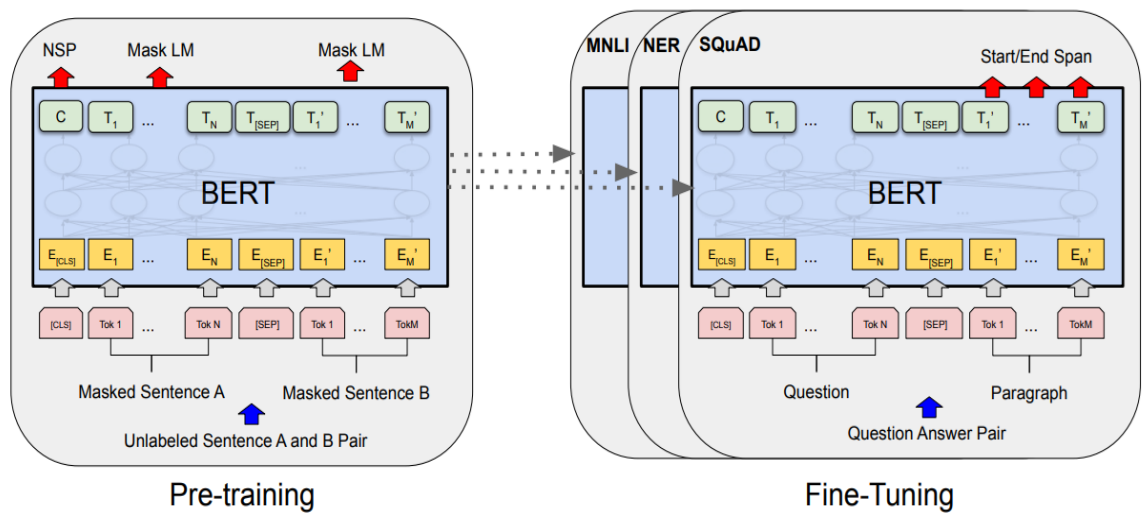

Figure 6. BERT's pre-training and fine-tuning procedures. Apart from output layers, the same architecture is used in both pre-training and fine-tuning stages. The same pre-trained model parameters are used to initialize models for different downstream tasks. During fine-tuning, all parameters are fine-tuned. [CLS] is a special symbol added at the beginning of every input sequence to represent sentence-level classification, and [SEP] is a unique separator token to separate two sequences, e.g., questions from answers [21].

BERT can be used for various language tasks, such as sentence classification, Question Answering (QA), and Named Entity Recognition (NER) with finetuning and minor modifications to the original architecture.

## 4.2 Vision Transformer (ViT)

ViT is a pure Transformer architecture without convolutional layers and was proposed for image classification tasks [1]. Like BERT, ViT is also an encoder-only Transformer model. Transformers cannot directly process spatial data such as images; therefore, data must be converted to a sequence. ViT splits an image into fixed-size patches, generally 16×16 or 32×32 flattened, before they are provided as an input to the transformer model, as shown in Fig 7. The flattened patches are placed in a sequence, then transformed into a low-dimensional linear embedding. Like the original Transformer, PEs are added to the linear embeddings to inject information about each patch's relative location in the image, where 1D, 2D, and learnable positional embedding can be used. An extra learnable class embedding is added at the start of the sequence, used for downstream classification tasks. During fine-tuning, a classification head comprised of a single hidden layer network is attached to this class embedding.

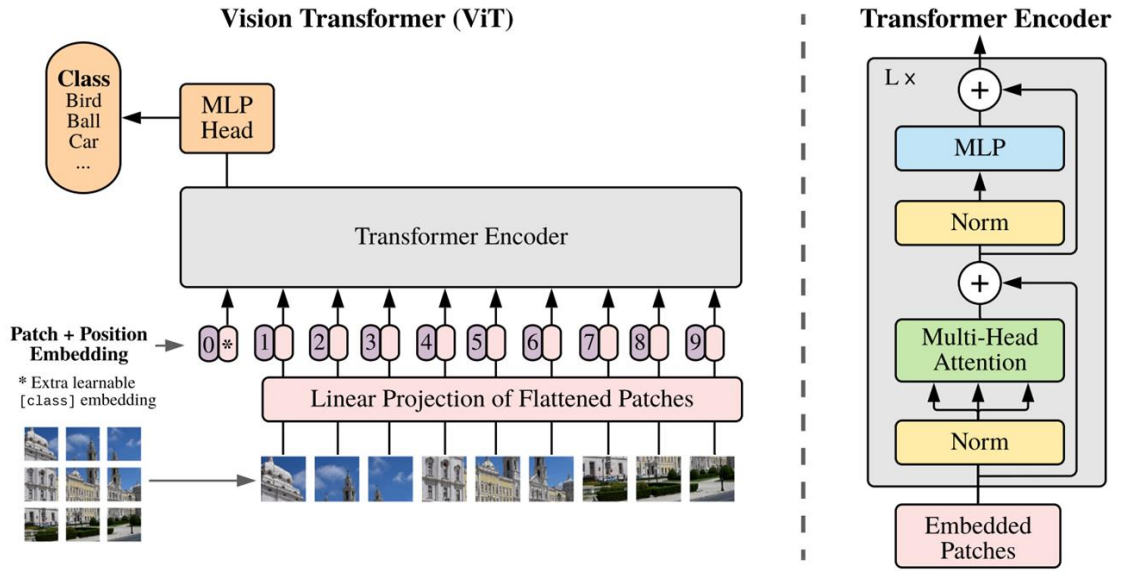

Figure 7. ViT splits an image into fixed-size patches, then linearly embeds the patches, adds position embeddings, and feeds the resulting sequence of vectors to a standard Transformer encoder. To perform classification, one would use the standard approach of adding an extra learnable "classification token" to the sequence ("CLS") [22].

Transformers models by design do not possess the inductive biases of CNNs, such as limited receptive field and translational invariance (ability to detect or recognize an object regardless of its location in an image). In CNNs, the receptive field increases linearly with the depth of the model. While the Transformer lacks the inductive biases of the CNN, they are permutation invariant (not dependent on the order of elements in a sequence), and the shallow layers of the model can attend to the entire image.

## 5   Large Language Models (LLMs)

Foundation models are large-scale AI systems trained on vast amounts of data to be adapted for a wide range of downstream tasks [23]. LLMs colloquially refer to a class of foundation models with parameters on the order of billions trained on language corpora with billions of words to generate human-like language and solve different NLP tasks. Most LLMs use the Transformer architecture, the current default architecture for processing sequential data as of 2023. The success of LLMs comes from the self-supervised pre-training paradigm, which takes advantage of large free text data without annotation. This pre-training technique enabled LLMs to generate coherent and realistic language, making them useful for various applications such as text completion, dialogue generation, and content generation. Large generative AI models trained to generate text and question answering are autoregressive decoder-only language models. Examples of autoregressive decoder-only language models include PaLM [24], GPT-3 [25], Chinchilla, LLaMA [26], PaLM2 [27] used in BARD chatbot, and GPT-4 [28]. These models are trained on billions of tokens obtained from datasets such as Common Crawl, WebText2, Books1, Books2, Wikipedia, Stack Exchange, PubMed, ArXiv, Github, Gutenberg, and many more. Some of the domain-specific LLMs include Galactica [29], trained on curated human scientific knowledge corpora, BloombergGPT [30], trained on proprietary financial data, and CodeX [31] for code generation. A timeline of popular LLMs is displayed in Fig. 8.

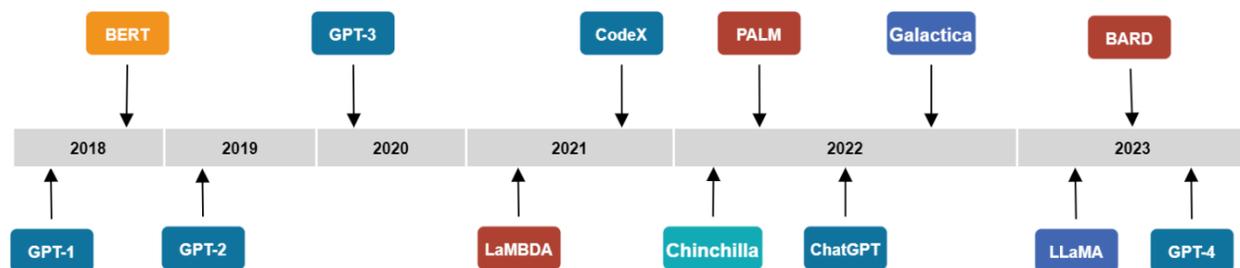

Figure 8. The timeline of popular large language models developed over the years (2018-2023).

The number of parameters in LLMs and the size of their training data has increased rapidly, reaching up to trillions of tokens [26]. The capabilities of LLMs appear to be a function of the amount of data, parameters, and computation resources rather than architectural design advancements [32]. The scaled-up language models develop abilities beyond the trained outcomes called 'emergent abilities,' which are not designed but discovered after deployment [33]. For example, GPT-3 showed few-shot prompting ability; when provided few input-outputs for a natural language task, the model can perform the task on unseen samples without further training or gradient updates to the parameters [25]. Parameter-efficient models such as Stanford Alpaca [32] and efficient finetuning approaches of Quantized LLMs such as QLoRA [34] have been introduced to address situations where computational resources are limited. Despite the exceptional ability of LLMs to generate realistic text, they also generated false information, toxic language, and racial stereotypes [35, 36].

In the medical domain, Agrawal et al. [37] demonstrated that LLMs can be few-shot clinical information extractors without further training on the clinical data. They used InstructGPT [38] for this task, significantly outperforming existing zero-shot and few-shot baselines. In Radiology, Jeblick et al. [39] performed an exploratory case study to evaluate ChatGPT's ability to simplify radiology reports. Expert human radiologists considered the simplified reports complete, factual, and devoid of harmful text that could misguide the patient. However, instances of missing key findings and incorrect statements were observed. The PMC-LLaMA [40] model, fine-tuned on 4.8 million biomedical papers obtained from PubMed Central, demonstrated a better understanding of biomedical domain-specific concepts than the original LLaMa when evaluated on biomedical QA benchmarks. GatorTron [41], a large clinical language model with 8.9 billion parameters trained on over 90 billion words of clinical text, was applied to clinical NLP tasks such as clinical concept extraction. Luo et al. [42] proposed BioGPT, a biomedical domain specific generative model pretrained on PubMed abstract corpus to generate fluent biomedical term descriptions.

Singhal et al. [43] evaluated the 540 billion parameters PaLM [24] and its variant FLAN-PaLM [44] on the benchmark dataset MultiMedQA. This benchmark dataset combines multiple QA datasets, including medical exams, consumer queries, and research. The authors also introduced Med-PaLM, a parameter-efficient model that used prompt instruction tuning to fix the critical Flan-PaLM gaps observed upon human evaluation. In subsequent work, Singhal et al. proposed Med-PaLM2 [45] to bridge the gap between the model's answers to that of clinicians. The model combines improvements that come with PaLM2 [27], a novel ensemble refinement prompting strategy, and domain-specific model finetuning. Scaled-up models such as ChatGPT, PaLM, PALM2, and GPT-4 have been shown to answer medical questions and successfully pass or achieve near-passing scores on medical licensing examinations [43, 46-49].

The impressive advancements of foundation models have not yet permeated into medical AI. These early approaches are limited by a lack of large, diverse medical datasets, the complex nature of medical data, federal patient data privacy regulations, and the recency of the general-purpose foundation models [50].

# 6 Transformers in NLP
## 6.1 Clinical Word Embeddings

Word embeddings map variable-length words to a fixed-length vector while preserving syntactic and semantic information. Word embeddings are a standard representation used in NLP. Traditional word embedding techniques such as word2vec [51] or GLoVe [52] learn an aggregated representation of all contexts associated with a word. Previously contextual word embedding based on models such as ELMo [53], BERT [21], and ULMFiT [54] achieved SOTA performance on NLP tasks. However, these embeddings cannot be adapted directly to clinical or biomedical text due to differences in the linguistic domain corpora. Lee et al. [50] introduced BioBERT, a pre-trained language model in the biomedical domain, to overcome this difficulty. BioBERT is initialized with BERT weights and is pre-trained on PubMed central full-text articles and abstracts as shown in Fig 9. This pre-trained model is fine-tuned on three popular biomedical NLP tasks, NER, Relation Extraction (RE), and QA. BioBERT has outperformed previous models on biomedical text mining tasks with minimal task-specific modification.

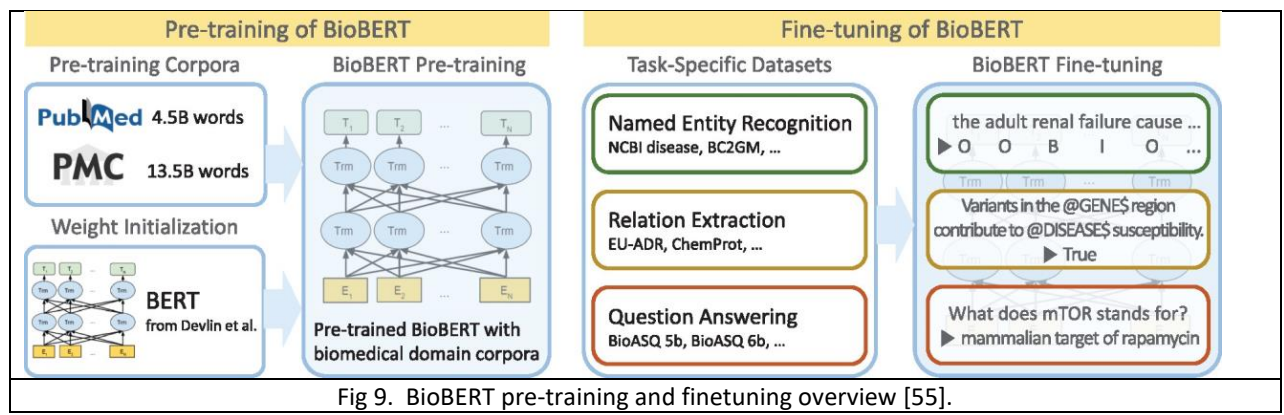

Fig 9. BioBERT pre-training and finetuning overview [55].

Further specialization of BERT and BioBERT via pre-training on specific EHR databases has proven promising. Alsentzer et al. [56] pre-trained BERT and BioBERT on 2 million clinical notes from the MIMIC-III database [57] to obtain clinical BERT and Bio+clinical BERT. Si et al. [58] explored various embedding methods such as word2vec [51], GloVe [52], fastText [59], ELMo [53], and BERT [21] on clinical concept extraction tasks to demonstrate the generalizability of these traditional embedding methods. When pretrained on a clinical domain-specific corpus [57], all the embeddings yielded increased performance. Huang et al. [60] pretrained BERT [21] on clinical notes from the MIMIC-III dataset [57] to develop ClinicalBERT. ClinicalBERT achieved higher Pearson correlation scores than word2vec [51] and fastText [59]. All these models were pre-trained on clinical domain corpora and have outperformed models pre-trained on general or biomedical domain corpora in clinical NLP tasks.

## 6.2 Transformers for Clinical Information Extraction (IE)

EHRs contain a wealth of patient information stored in structured and unstructured formats, including detailed clinical notes used for documentation. Parsing through this data is difficult due to the unstructured nature of the free text entries recorded by clinical staff in the EHR. Clinical IE consists of sub-tasks such as NER, coreference resolution (CR), QA, semantic textual similarity (STS), relationship extraction (RE), and entity normalization (EN). The success of Transformers inspired researchers to adapt Transformer-based architectures for clinical IE. Table 2 shows a list of Transformer based language models in clinical and biomedical domains, the NLP tasks performed using the models, and datasets used.

| Table 2. Transformers in Clinical and Biomedical NLP | | | | |
|---|---|---|---|---|
| NER: Named Entity Recognition; SS: Sentence Similarity; RE: Relation Extraction; DC: Document Classification; NLI: Natural Language Inference; QA: Question Answering; EN: Entity Normalization; STS: Semantic Textual Similarity | | | | |
| Reference | Title | Tasks | Datasets | Architecture |
| [55] | BioBERT: a pre-trained biomedical language representation model for biomedical text mining | NER, Relation extraction, Question answering | NCBI Disease [61], I2b2 2010 [62], BC5CDR [63], BC4CHEMD [64], BC2GM [65], JNLPBA [66], LINNAEUS [67], Species-800 [68], GAD [69], EU-ADR [70], CHEMPROT [71], BioASQ [72] | BERT[21] |
| [56] | Publicly available clinical BERT embeddings | NLI, NER, de-identification, concept extraction, entity extraction | MIMIC-III [57], i2b2 2010 [62], i2b2 2012 [73, 74], MedNLI [75], i2b2 2006 [76], i2b2 2014 [77, 78] | BERT [21] |
| [58] | Enhancing clinical concept extraction with contextual embeddings | Concept extraction | i2b2 2010 [62], i2b2 2012 [73], i2b2 2014 [77], ShARe/CLEF [79, 80], SemEval [81-83], MIMIC-III [57] | BERT [21] |
| [84] | BlueBERT: Transfer Learning in Biomedical Natural Language | SS, NER, RE, DC, Inference | MEDSTS [85], BIOSSES [86], BC5CDR [63], | BERT[21] |

| | Processing: An Evaluation of BERT and ELMo on Ten Benchmarking Datasets | | ShARe/CLEF [79], DDI [87], CHEMPROT [71], i2b2 2010 [62], HoC [88], MedNLI [75] | |
|---|---|---|---|---|
| [89] | Domain-Specific Language Model Pretraining for Biomedical Natural Language Processing | NER, RE, SS, DC, QA | NCBI Disease [61], BC5CDR [63], BC2GM [65], JNLPBA [90], CHEMPROT [71], DDI [87], GAD [69], BIOSSES [86], HoC [88], PubMedQA [91] BioASQ [72, 92] | PubMedBERT |
| [60] | ClinicalBERT: Modeling Clinical Notes and Predicting Hospital Readmission | Patient readmission prediction | MIMIC-III [57] | BERT [21] |
| [93] | Clinical concept extraction using transformers | Concept extraction | MIMIC-III [57], i2b2 2010 [62], i2b2 2012 [73, 74], n2c2 2018 [94, 95] | BERT [21], RoBERTa [96], ALBERT [97], ELECTRA [98] |
| [99] | Relation Extraction from Clinical Narratives Using Pre-trained Language Models | Relation extraction | n2c2 2018 [94, 95]. i2b2 2010 [62] | BERT [21] |
| [100] | Transformer-Based Argument Mining for Healthcare Applications | Argument component detection, Relationship classification | MEDLINE | BERT [21], BioBERT [55], SciBERT [101], RoBERTA [96] |
| [102] | Clinical XLNet: Modeling Sequential Clinical Notes and Predicting Prolonged Mechanical Ventilation | Prognosis prediction | MIMIC III [57] | XLNet [103], BERT [21], ClinicalBERT [60], |
| [104] | BioBERT based named entity recognition in electronic medical record | NER | I2b2 2010 [62] | BioBERT[55] |
| [105] | Multiple features for clinical relation extraction: A machine learning approach | Relation extraction | n2c2 2018 [94, 95], MADE 2018 [106] | BERT [21], BioBERT [55], ClinicalBERT [60] |
| [91] | PubMedQA: A dataset for biomedical | QA | PubMedQA [91] | BioBERT [55] |

| | research question answering | | | |
|---|---|---|---|---|
| [107] | Pre-trained language model for biomedical question answering | QA | SQuAD [108, 109], BioASQ [72, 92] | BioBERT [55] |
| [110] | BERT-based ranking for biomedical entity normalization | EN | ShARe/CLEF [111], NCBI [61], TAC2017ADR[112] | BERT [21], Bio BERT [55], ClinicalBERT [56] |
| [113] | Measurement of Semantic Textual Similarity in Clinical Texts: Comparison of Transformer-Based Models | STS | 2019 n2c2/Open Health NLP [114] | BERT [21], XLnet [103], RoBERTa [96] |

### 6.2.1 Named Entity Recognition

Clinical named entity recognition (CNER) aims to identify entities, concepts, and events such as disease, drugs, treatments, medical conditions, and symptoms from clinical narratives. CNER is challenging as clinicians often use acronyms and abbreviations to describe complex clinical terms without using standardized clinical ontology. Earlier approaches used the BERT model to generate clinical textual embeddings, which were further used to train other deep learning models, such as Bi-LSTM and conditional random fields [115-117]. Later, for biomedical and clinical domains, domain-specific BERT-based models such as BioBERT [55] and clinical BERT by Alsentzer et al. [56] established baselines on CNER datasets. BERT-based models have been applied to CNER tasks in different languages, such as Chinese [118, 119], Korean [120], Italian [121], Spanish [122], and Arabic [123].

The clinical de-identification task, which removes protected health information, was also approached as a NER problem by pretrained BERT-based models, such as clinical-BERT [56] and UMLS-BERT [124]. These models were applied to i2b2-2006 [76] and i2b2-2014 [78] de-identification tasks. Garcia et al. [125] and Mao et al. [126] used BERT on the MEDDOCAN [127] Spanish de-identification corpus.

The clinical concept extraction task predicts a concept's start and end positions in a document. BIO tags are commonly used, where "B", "I", and "O" refer to the beginning, inside, and outside of a concept. Yang et al. [93] developed an open-source Transformers package with four transformer-based models, BERT [21], ALBERT [97], RoBERTa [96], and ELECTRA [98], pretrained on MIMIC-III dataset for clinical concept extraction. Peng et al. [84] used transfer learning to fine-tune BERT [21] for concept extraction on BC5CDR [63] and ShARe/CLEF [111] datasets. Khan et al. [128] proposed MT-BioNER, a transformer-based model for intent classification and slot tagging. The authors combined BERT encoder layers with task-specific layers to train their model on NCBI-disease [129], BC5CDR [63], and JNLPBA [90] datasets.

### 6.2.2 Clinical Coreference Resolution (CR)

The CR task aims to identify all mentions of the same entity in a text. Trieu et al. [130] performed CR in full-text articles as part of the CRAFT 2019 shared task [131]. The authors employed a span-based end-to-model proposed by Lee et al.[132] and replaced the LSTM layers with BERT. Their results on the CRAFT coreference resolution task indicate the effectiveness of BERT in capturing long-distance coreferences in large documents. Steinkamp et al. [133] used BERT [21] to perform CR for symptom extraction on the i2b2 2009 Medication Challenge [134] and MIMIC-III datasets [57], showing better performance compared to recurrent models.

### 6.2.3 Clinical Relationship Extraction (CRE)

CRE is categorized into concept relationship and temporal relationship extraction. Concept relationship extraction identifies the relationship between two concepts (e.g., drug and dosage), whereas temporal relationship extraction evaluates the relationship between clinical events occurring at different times. Peng et al. [84] approached the CRE task as a sentence classification problem by replacing named entity mentions of interest with pre-defined tags using BERT [21] on DDI [87], ChemProt [71], and i2b2 2010 [62] datasets. Wei et al. [99] fine-tuned BERT outperformed SOTA RE models on clinical RE tasks using n2c2-2018 [95] and i2b2-2010 [62] datasets. Zhang et al. [115] pretrained the BERT model on Chinese clinical text and fine-tuned on the breast cancer dataset to classify the relationship between clinical concepts and corresponding attributes for breast cancer. Using BERT, Xue et al. [129] used an integrated joint learning approach for NER and CRE in coronary angiography Chinese clinical text. Lai et al. [135] proposed BERT-GT, which combines BERT with Graph Transformer by integrating the neighbor attention mechanism into BERT. BERT-GT was used for cross-sentence RE on the N-ary [136] and BioCreative CDR [137] datasets. Lin et al. [138] developed a pre-trained BERT model on the MIMIC-III dataset and BioBERT [55] models for temporal RE on the THYME [139] corpus. Their BioBERT model with sentence agnostic 60-token window approach was used for the CONTAINS temporal relation extraction task on the colon cancer test set.

### 6.2.4 Question Answering (QA)

The (QA) ability of a model can serve as an indicator of its ability to learn the medical text. Jin et al. [91] introduced the PubMedQA dataset for biomedical research question answering, and fine-tuned BioBERT model to establish a baseline on the dataset. Yoon et al. [107] pretrained the BioBERT model on SQuAD [108, 109] datasets and fine-tuned it for the BioASQ [72, 92] biomedical QA challenge. This model achieved SOTA performance on factoid, list, and yes/no type questions of the BioASQ dataset. He et al. [140] proposed a procedure for consumer health question answering and medical language inference tasks using models such as BERT[21], BioBERT[55], SciBERT[101], ClinicalBERT[56], BlueBERT[84], and ALBERT[97]. Schmidt et al. [141] developed a QA-BERT model for question answering using the PICO (Population, Intervention, Comparator, and Outcome) framework. The PICO element dataset [142] was combined with SQuAD datasets [108, 109] to increase the generalizability and flexibility of the model on all types of questions. The proposed QA-BERT performed better than LSTM and BERT baselines [141].

### 6.2.5 Biomedical Entity Normalization (BEN)

BEN aims to link mentions of an entity in a clinical document (e.g., EHR) to their corresponding concepts in a knowledge base [143]. Ji et al. [110] fine-tuned pre-trained models such as BERT [21], BioBERT [55], ClinicalBERT [56] on three different datasets ShARe/CLEF [111], NCBI [61], TAC2017ADR[112] for performing BEN. Li et al. [144] proposed the EhrBERT model, pre-trained on 1.5 million EHR notes, and evaluated it on three entity normalization corpora, namely the MADE corpus [106], NCBI disease corpus [61], and CDR corpus [63]. Authors observed that their models performed worse when the pre-training domain and fine-tuning task were distant.

### 6.2.6 Semantic Text Similarity (STS)

STS is an NLP task that measures the similarity between two pieces of text using a pre-defined metric. Xiong et al. [145] proposed a gated network to fuse one hot and distributed representations obtained from sentence-level features like inverse document frequency, sentence length, N-gram overlaps, and similarity metrics between two input sentences. Their fusion-gated BERT model was used on the clinical STS task of the BioCreative/OHNLP 2018 challenge [146]. Yang et al. [113] explored three models, BERT [21], XLnet [103], and RoBERTa [96], for clinical STS as a part of the 2019 n2c2/Open Health NLP challenge [114]. The Models were pre-trained on a general STS dataset and fine-tuned on the clinical STS training partition. Among these, RoBERTa-large achieved the highest performance.

### 6.2.7 Automatic International Statistical Classification of Diseases (ICD) Coding

ICD codes are a set of alpha-numeric designations to communicate diseases, symptoms, procedures, diagnoses, and abnormal findings in a universally accepted way among healthcare professionals. ICD coding involves recording the ICD codes associated with a patient's visit. This coding process is often performed manually, which may result in documentation errors and consume a significant amount of time. Zhang et al. [147] proposed BERT-XML with multi-label attention to model 2292 ICD-10 codes from EHR notes [148]. Biswas et al. [149] used a transformer-based encoder architecture TransICD with a structured self-attention mechanism [150] to extract label-specific representations for multi-label ICD coding. Label distribution aware margin loss [151] was used to address the imbalance in ICD codes data. Transformer-based automatic ICD coding was used in clinical texts of Chinese [152], Spanish [153, 154], Swedish [155], and Thai [156]. Silvestri et al. [157] used a Transformer Cross-lingual Language Model(XLM) [158] for automatic ICD coding by fine-tuning clinical texts in English and testing on clinical Italian text.

## 6.3 Neural Machine Translation(NMT)

Automatic NMT of biomedical data is essential to make healthcare information available to medical professionals and general public to overcome language barriers. Tubay et al. [159] for the low-resourced biomedical NMT task used a Transformer model enhanced with multi source translation technique capable of exploiting multiple text inputs from the same language family. Berard et al. [160] proposed a multilingual neural machine translation(MNMT) model to translate biomedical text from 5 different languages French, Spanish, German, Italian, and Korean to English. The MNMT model is a variant of Transformer Big architecture with complex encoder capable of representing multiple languages. Liu et al. [161] proposed BioNMT Transformer model to translate domain specific biomedical vocabulary from foreign languages. The model is capable of semantic disambiguation of unknown words in the translation using external biomedical dictionaries to replace the unknown words. Wang et al. [162] used Transformer large model with 20 encoder layers for biomedical translation shared task to translate German, French, and Spanish to English at Workshop on Machine Translation. Subramanian et al. [163] used Transformer model for the same biomedical shared task at WMT to translate text from English to German and Russian. Their transformer model used a combination of model scaling, data augmentation with back-translation, knowledge distillation, model ensembling, and noisy channel re-ranking to perform the translation task.

## 7 Transformers for Structured EHR Data

Structured EHR data includes ICD codes for diagnoses, medication, age, and other demographics collected every time a patient visits the hospital. These data are linked by an underlying temporal structure representing the cycle of diagnosis, medication/intervention, and potential patient readmissions. Furthermore, medication and diagnosis codes are derived from an ontological tree structure. Therefore, clinical tasks such as predicting future disease diagnoses, readmissions, or mortality rely on accurately representing the temporal and graphical structure of a patient's EHRs. This challenge has led to three broad NLP tasks on structured EHR content that have been attempted in recent years using transformer networks.

### 7.1 Ontological Structure Learning

Previous studies have tried to learn the graphical structure inherent within the EHR using novel Transformer architectures. Choi et al. proposed the Graph Convolution Transformer (GCT) to jointly learn the relationships between diagnoses and medication codes while performing diagnosis-treatment classification [164]. They used conditional probabilities between medications and diagnoses calculated over the entire dataset to guide the attention maps in their Transformer network. Their model was validated on the eICU collaborative research dataset [165]. In contrast, Shang et al., 2019 explicitly used graph neural networks (GNN) for learning medical ontology embeddings and used these embeddings in a transformer to recommend future medications using the MIMIC-III dataset [166]. To leverage the entire dataset, they pre-trained G-BERT, a combination of GNN and BERT, on EHR data with only one admission. Peng et al., used a graph-based attention model (GRAM) to create ontological embeddings, which were then represented using muti-head self-attention to learn the ontological structure of medications within EHR [167].

### 7.2 Multi-modal Data Fusion

Previous studies have used Transformer network to create joint embeddings amongst multiple data modalities, such as EHR and clinical notes. Darabi et al., 2020 used separate Transformer networks to create different representations for the clinical codes (ICD, drug, and procedure) and clinical notes and combined them into one "patient representation" [168]. They used this joint representation to predict future diagnoses, procedures, length of stay (LOS), readmission, and mortality. Studies have used joint-embeddings in BERT to predict rare diseases such as chronic cough [169] or depression [170]. Xu et al., 2021 proposed the use of multi-modal fusion architecture search (MUFASA), using an evolutionary algorithm to jointly search for the optimal architecture to represent subsets of EHR data and the optimal stage at which the individual embeddings will undergo fusion [171]. In contrast, Zhang et al., 2021 used a contrastive learning approach to increase the mutual agreement between different modalities for the same patient and increase the contrast for the same modality amongst different patients while jointly optimizing a prediction loss [172]. They showed that combining this representation with the BERT encoder predicted mortality and length of stay better than other baselines.

## 7.3 Predicting Future Diagnoses using ICD Codes

BEHRT, an adaptation of BERT on EHR data, was trained from scratch using the masked language modeling task on sequential ICD codes and age to predict future diagnoses [173]. This model was developed primarily on the UK Clinical Practice and Research Datalink (CPRD) [174]. Recently, BEHRT was used to predict incident heart failure [175] and to perform causal inference [176]. The Hi-BEHRT model extended this by incorporating self-supervised pretraining by masking certain EHR data and certain time points in patients' visitation history and creating localized feature aggregator Transformer embeddings fused at a later stage using global attention [177]. Hi-BEHRT performed better than BEHRT in predicting the onset of heart failure, diabetes, chronic kidney disease, and stroke. Compared to the BEHRT-based models, Med-BERT expanded the pretraining task to include prediction of prolonged length of stay and used a combination of ICD-9 and ICD-10 codes to create their model, which was subsequently evaluated on predicting diabetes and heart failure [178]. Another model, HiTANet, explicitly included a time vector to represent the time elapsed between consecutive visits. The time embedding was combined with the original visit embedding and used as key values in a global attention block to represent the most significant time points in a patient's medical history [179]. They tested their model efficacy in predicting future diagnoses of three disease-specific datasets. The RAPT model combined an explicit time-span information vector with additional pre-training tasks such as similarity prediction and reasonability check to address data insufficiency, incompleteness, and short sequence problems inherent in EHR data [180]. They evaluated their model for predicting pregnancy outcome, risk period, and the diagnoses of diabetes and hypertension during pregnancy.

# 8 Transformers in Computer Vision

## 8.1 Medical Image Segmentation

Image segmentation is a dense pixel classification task which requires capturing the complex interactions between individual pixels of an image. Unlike general purpose image segmentation, medical image segmentation suffers from a lack of large datasets, requires the context of surrounding anatomical structures, and must account for inter-patient anatomical variabilities. Several data modalities, such as X-ray, ultrasound, magnetic resonance imaging (MRI), computed tomography (CT), positron emission tomography (PET), and microscopy can benefit from medical image segmentation. Prior to the success of Transformer models, the U-net architecture, proposed by Ronneberger et al. [181], was the prominent architecture for medical image segmentation. The U-net model is a Convolutional Neural Network (CNN). Convolutional layers are limited in long-range feature modeling. This is because the receptive field of convolutional filters increases linearly and therefore only the deepest convolutional layers have the global context of an image. Although incorporating dilation and stride into convolution can address the limitations of long-range dependencies to some extent, it results in an unavoidable tradeoff between global and local information. On the contrary, the self-attention mechanism in Transformer layers can model the global context of images, irrespective of its depth in the network.

Researchers have used transformer-based models to segment different tissues and organs such as heart, abdominal organ, brain tissue, skin lesion, prostate, gland, polyp, hip, thoracic, and lung segmentation. A comprehensive list of transformer-based models used for performing the above-mentioned segmentation objectives is provided in Table 3. Medical images from different modalities can come in 2D or 3D formats.

| Table 3. Transformers for Medical Image Segmentation | | | | |
|---|---|---|---|---|
| Reference | Title | Datasets | Task | Modalities |
| [182] | TransUNet: Transformers Make Strong Encoders for Medical Image Segmentation | Synapse [183], ACDC [184] | Multi-organ segmentation, Cardiac segmentation | CT, MRI |
| [185] | Medical Transformer: Gated Axial-Attention for Medical Image Segmentation | Brain Segmentation, GLAS [186], MoNuSeg [187, 188] | Brain-anatomy segmentation, Gland segmentation, Nucleus segmentation | Ultrasound, Microscopy |
| [189] | TRANSCLAW U-NET: CLAW U-NET WITH TRANSFORMERS FOR MEDICAL IMAGE SEGMENTATION | Synapse [183] | Multi-organ segmentation | CT |
| [190] | UNETR: Transformers for 3D Medical Image Segmentation | BCV [183], MSD [191] | | CT |
| [192] | UTNet: A Hybrid Transformer Architecture for Medical Image Segmentation | M&Ms [193] | Cardiac segmentation | MRI |
| [194] | TransFuse: Fusing Transformers and CNNs for Medical Image Segmentation | Kvasir [195], CVC-Clinic [196], CVC-Colon [197], EndoScene [198], ETIS [199], | Polyp segmentation, Skin lesion segmentation, Hip segmentation. Prostate segmentation | Colonoscopy, |
| [200] | CoTr: Efficiently Bridging CNN and Transformer for 3D Medical Image Segmentation | BCV [183] | Multi-organ segmentation | CT |
| [201] | Swin-Unet: Unet-like Pure Transformer for Medical Image Segmentation | Synapse [183], ACDC [184] | Multi-organ segmentation, Cardiac segmentation | CT MRI |
| [202] | MISSFormer: An Effective Medical Image Segmentation Transformer | Synapse [183], ACDC [184] | Multi-organ segmentation, Cardiac segmentation | CT MRI |

| Ref | Title | Datasets | Tasks | Modality |
|---|---|---|---|---|
| [203] | Pyramid Medical Transformer for Medical Image Segmentation | GLAS [186], MoNuSeg [188], HECKTOR [204] | Gland segmentation, Nucleus segmentation Tumor segmentation | Microscopic images, CT/PET |
| [205] | Multi-Compound Transformer for Accurate Biomedical Image Segmentation | Pannuke[206], CVC-Clinic [196], CVC-Colon [197], ETIS [199], Kvasir [195], ISIC2018 [207] | Cell segmentation, Polyp segmentation, Skin lesion segmentation | Pathology, Colonoscopy, Dermoscopy |
| [208] | DS-TransUNet: Dual Swin Transformer U-Net for Medical Image Segmentation | CVC-Clinic [196], CVC-Colon [197], EndoScene [198], ETIS [199], GLAS [186], Kvasir [195], ISIC2018 [207] | Polyp segmentation, Skin lesion segmentation, Gland segmentation, Nucleus segmentation | Pathology, Colonoscopy, Dermoscopy |
| [209] | Medical Image Segmentation Using Squeeze-and-Expansion Transformers | REFUGE2020 [210], Drishti-GS [211], RIM-ONE v3 [212], Kvasir [195] | Optic disc and cup segmentation, Polyp segmentation, Brain tumor segmentation | Colonoscopy, MRI, Fundus images |
| [213] | SpecTr: Spectral Transformer for Hyperspectral Pathology Image Segmentation | Choledoch database [214] | | Pathology |
| [215] | LeViT-UNet: Make Faster Encoders with Transformer for Medical Image Segmentation | Synapse [183], ACDC [184] | Multi-organ segmentation, Cardiac segmentation | CT MRI |
| [216] | Transbts: Multimodal brain tumor segmentation using transformer | BraTS 2019 [217, 218], BraTS 2020 [217, 218] | Brain tumor segmentation | MRI |
| [219] | TransAttUnet: Multi-level Attention-guided U-Net with Transformer for Medical Image Segmentation | ISIC 2018 [207], JSRT[220], Montogomery [221], NIH [222], Clean-CC-CCII [223], GLAS [186], Bowl [224] | Chest X-ray segmentation, Skin lesion segmentation, Nucleus segmentation, Gland segmentation | X-ray, Histology, CT |
| [225] | U-net transformer: self and cross attention for medical image segmentation | TCIA, Internal dataset | Abdominal organ segmentation | CT |
| [226] | AFTer-UNet: Axial Fusion Transformer UNet for Medical Image Segmentation | BCV [183], Thorax-85 [227], Segthor [228] | Multi-organ segmentation, Thoracic segmentation | CT |
| [229] | A Transformer-Based | MSD [191] | | CT |

| | | | |
|---|---|---|---|
| | Network for Anisotropic 3D Medical Image Segmentation | | |
| [230] | HybridCTrm: Bridging CNN and Transformer for Multimodal Brain Image Segmentation | MRBrainS [231], iSEG-2017 [232] | Brain tissue segmentation, | MRI |
| [233] | Self-Supervised Pre-Training of Swin Transformers for 3D Medical Image Analysis | BTCV [183], MSD [234] | Multi-organ abdominal segmentation | CT |
| [235] | TiM-Net: Transformer in M-Net for Retinal Vessel Segmentation | STARE [236], CHASEDBI [237], DRIVE [238] | Retinal vessel segmentation | Color images |
| [239] | Auxiliary Segmentation Method of Osteosarcoma in MRI Images Based on Denoising and Local Enhancement | | Osteosarcoma segmentation | MRI |
| [240] | Dilated transformer: residual axial attention for breast ultrasound image segmentation | BUSIS [241] | Breast segmentation | Ultrasound |
| [242] | ColonFormer: An Efficient Transformer Based Method for Colon Polyp Segmentation | Kvasir [195], CVC-Clinic[196], CVC-Colon [197], CVC-T [198], ETIS [199] | Polyp segmentation | Colonoscopy |

### 8.1.1 CNN-Transformer Hybrids

The majority of approaches for transformer-based medical image segmentation used Transformers in conjunction with U-Net [181]. TransUNet, proposed by Chen et al. [182], is shown in Fig 10 and was one of the earliest examples. TransUNet uses a CNN to downsample the input image before providing it to a Transformer encoder which creates a global contextualized deep representation of the image. This representation is subsequently passed through a cascaded up-sampler to convert it into the full-resolution segmented output image. CNN downsampling is used to reduce the computational complexity of TransUNet architecture. This idea of using a Transformer as an U-net encoder to learn long range dependencies was subsequently adapted by multiple studies such as TransClaw U-Net [189], BiTr-UNet [243], Bi-FPN-UNet [244], and Weaving Attention U-Net [245]. UNet-Transformer used MHSA in skip-connections between the encoder and the decoder to recover finer spatial features [225]. LeViT-Unet [215] integrated LeVIT [246] into the downsampling block of U-net. TransAttUnet [219] used a novel self-aware attention module with both Transformer self-attention and global spatial attention.

In the domain of 3D medical image segmentation, UNETR [247] used ViT-B16 [248] as the encoder instead of CNN while retaining the U-shaped network design. TransBTS used 3D CNN blocks as the encoder to model spatial information followed by Transformer encoder to capture long distance dependencies and decoder to model volumetric data in MRI scans [216]. CoTr concatenated CNN feature maps at different scales using positional encoding and passed them into stacked Deformable Transformer encoder blocks [249]. Deformable Transformer computed attention over a local region around reference points instead of global self-attention, thereby reducing the computational complexity. The authors showed that this methodology out-performed other CNN-Transformer hybrid models on the BCV dataset [183] that covers 11 major human organs. SpecTr [213] used adaptively sparse Transformer blocks [250] to remove redundant/noisy bands of spectral information in the Transformer encoder during segmentation of hyperspectral images. This study also used 3D CNN encoders in combination with Transformer encoders in a U-Net fashion. nnFormer [251] is a 3D Transformer for volumetric image segmentation that used interleaved convolutional and local/global self-attention operations coupled with skip attention between the encoder and decoder to achieve better performance over other CNN-transformer hybrid models in three datasets [183, 184, 234]. Tang et al. [233] developed a new 3D Transformer-based model named Swin UNEt Transformer (Swin UNETR) with a hierarchal encoder for self-supervised pre-training using five public CT datasets. The model contains a Swin Transformer encoder that directly utilizes 3D patches and is connected to a CNN-based decoder via skip connections at different resolutions. The model was fine-tuned and validated using the BCV dataset [183] and the Medical Segmentation Decathlon (MSD) dataset [234]. These studies reflect the intention to find effective ways of combining convolutions with attention in medical image segmentation.

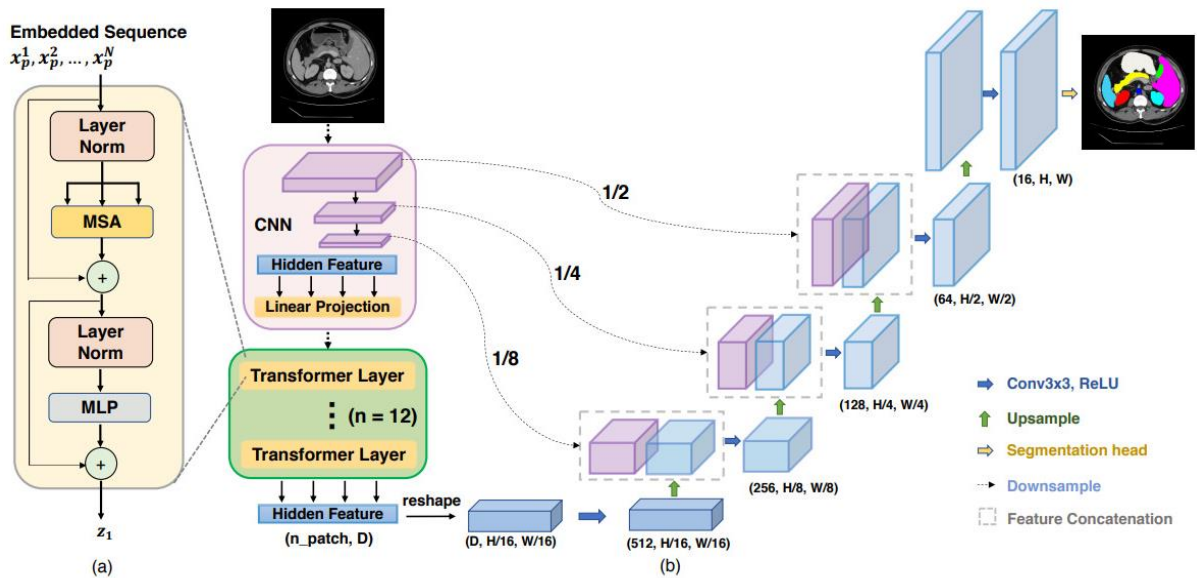

Figure 10: Overview of TransUNet architecture a) Schematic of Transformer encoder b) TransUNet architecture. The figure was adapted without modifications from [182].

### 8.1.2 Transformer-Only U-Nets

UTNet [192] introduced Transformer self-attention into both the encoder and decoder to capture long-range dependencies at different scales. Swin-Unet [201] used pure Swin Transformer [252] blocks. DS-TransUNet used a dual-branch Swin Transformer in the encoder to extract feature representations at multiple scales and Transformer Interactive Fusion (TIF) blocks to establish global interactions between them [208]. They also employed Swin Transformer blocks in both the encoder and decoder. Valanarasu et al. [185] proposed Medical Transformer (MedT) with a gated axial attention layer along with local and global branches (LoGo). The proposed gated axial attention layer was adapted based on position-sensitive axial attention [253] to influence positional bias on small-scale medical datasets. Karimi et al., 2022 developed a convolution-free 3D segmentation framework using pre-trained vanilla Transformer encoder which performed better than CNN models on three proprietary datasets [254].

### 8.1.3 Non U-Net Transformer Models

Zhang et al. [235] developed the TiM-Net model based on M-Net [255] with diverse attention mechanisms, and weighted side output layers for retinal vessel segmentation. The model was validated on three public retinal image datasets: STARE [236], CHASEDBI [237], and DRIVE [238]. Wang et al. [239] proposed an auxiliary segmentation method for osteosarcoma detection in MRI images based on denoising and local enhancement. For noise removal, the authors used the Eformer [256]. Duc et al. [242] developed a network called ColonFormer for polyp segmentation from endoscopic images on Kvasir [195], CVC-Clinic DB [196], CVC-Colon DB [197], CVC-T [198], and ETIS-Larib Polyp DB [199] datasets. The model uses Mix Transformer [257] as the encoder backbone, which is a hierarchical Transformer encoder that can represent both high and low resolution features. It also includes the efficient Self-Attention to reduce the computational complexity of self-attention layers.

## 8.2 Medical Image Registration

Image registration is the process of transforming data from multiple datasets into one coordinate system. Registration is essential for comparing, analyzing, or integrating data obtained from various sources, different viewpoints, different times, or different sensors [258]. Recent deep learning approaches have incorporated attention-based Transformer models for this task.

Chen et al. proposed one of the earliest Transformer based architectures, VIT-V-Net [259] which combines the vision Transformer (ViT) [248] and V-Net [260], a CNN architecture. Wang et al. [261] developed TUNet which incorporates ViT [22] into the U-Net [181] architecture to extract global and local features from the moving and fixed images to effectively generate the deformation field. Mok et al. [262] developed a fast and robust learning-based algorithm called C2FViT for 3D affine medical image registration. C2FViT leverages global connectivity and locality of the convolutional vision Transformer and a multi-resolution strategy to learn the global affine registration. Both the above papers evaluated their models on brain template-matching normalization and atlas-based registration using the OASIS [263] and LPBA [264] datasets. Tulder et al. proposed pixel and token wise cross-view attention to integrate multiple views in mammography and X-ray imaging [265] using CBIS-DDSM [266] and CheXpert [267] datasets.

Chen et al. proposed TransMorph [268], a modified U-net architecture that incorporates Swin Transformer [252] blocks in its down-sampling branch for unsupervised affine and deformable image registration on the IXI [269] dataset. Transformer blocks enabled the estimation of deformation uncertainty while preserving the registration performance. Zhu et al. [270] proposed the Swin-VoxelMorph, an unsupervised learning model which applies a hierarchical Swin Transformer [252] as the encoder to extract contextual information and a symmetric Swin Transformer-based decoder with a patch expanding layer to perform up-sampling to estimate the registration fields. The authors used two datasets to validate the model: ADNI [271] and PPMI [272].

### 8.3 Medical Image Captioning and Report Generation

Expert medical professionals typically interpret biomedical images, and their findings are documented as medical reports. Medical report writing is time-consuming and requires specialized personnel. Automated medical report generation can reduce the workload on doctors and reduce human errors.

Hou et al. [273] proposed the RATCHET model, a medical Transformer to generate medical text reports from chest X-rays. The authors used the MIMIC CXR v2.0.0 dataset [274] which has over 300,000 chest radiograph images and free-text radiology reports. Free text reports were tokenized using the byte pair encoding approach [275]. The RATCHET architecture follows the encoder-decoder architecture, but the encoder is a CNN model, DenseNet-121 [276], whereas the decoder is the vanilla Transformer decoder. The output features of the DenseNet-121 encoder are provided as input to the second attention block of the Transformer decoder, whereby the network learns context from the radiography image against the input text report. Free text tokens are shifted right and provided as input to the decoder to predict the next token. Nicholson et.al., 2021 used a pretrained ViT encoder and a pretrained PubMedBERT decoder to solve the ImageCLEFmed Caption task of 2021 [277]. Their model was fine-tuned on the ROCO dataset [278]. It was fine-tuned and tested on four datasets namely PadChest [279], CheXpert [267], ChestX-ray14 [280], and MURA [281] which is a musculoskeletal radiograph dataset.

Alfarghaly et al. [282] used conditioned self-attention, where new key and value parameters were introduced to project the encoder's output to the decoder's attention space. The authors used visual and semantic features extracted using Chexnet [283], a Densenet121 model and pre-trained word2vec embeddings, respectively. For the training and validation of the model, they used the IU-Xray dataset [284]. You et al. [285] developed an AlignTransformer for chest X-ray images consisting of two modules: Align Hierarchical Attention (AHA) and Multi-Grained Transformer (MGT). The AHA module was used to align visual regions and disease tags. Features from the AHA module were provided as input to the MGT module. The MGT module adaptively exploited multi-grained disease-grounded visual features to determine the importance of visual features for each target word. The authors used two publicly available datasets: IU-Xray [284] and MIMIC-CXR [286]. Pahwa et al. [287] developed a memory-driven

Transformer model called MedSkip for report generation. MedSkip consists of the standard Transformer encoder and a relational memory decoder. It was trained on Pathology Education Informational Resource (PEIR) Gross dataset [288] and IU X-Ray [284] datasets. Li et al. developed a Cross-modal clinical Graph Transformer (CGT) to incorporate expert knowledge into ophthalmic report generation [289]. The model first restores a sub-graph from the clinical graph and injects clinical relation triples into the visual features as prior knowledge. Finally, reports are predicted using the encoded cross-modal features using a Transformer decoder. The CGT model was trained and validated on an ophthalmic report generation dataset called FFA-IR [290].

## 8.4 Visual Question Answering (VQA)

VQA is a computer vision task where a question is posed and the answer must be inferred from an image. In the medical domain, VQA can be used to extract information from medical images to assist in making a diagnosis. Ren & Zhou, 2020 [291] developed the CGMVQA model, which modified the original Transformer by using layer normalization before the MHSA and FCFN layers. The model was trained and validated on the ImageCLEF 2019 VQA-Med data set [292]. The CGMVQA can interchangeably deploy a classification or a generative mode by changing the output layer and loss function while retaining the same architecture. While in the classification mode, the model can predict a yes-no, modality, plane or organ system answer, in the generative mode, the model uses masked answers to predict the next word in a sentence. Naseem et al. [293] introduced TraP-VQA model to answer medical questions presented in pathology images. This model embedded low-level visual features extracted using a CNN, low-level language features extracted using a domain-specific Language model and the Transformer layer to learn the contextualized representation between the two to solve the VQA task. The authors used the public PathVQA dataset [294] to train and validate their model. Sharma et al., 2021 [295] developed an attention-based multimodal deep learning model called *MedFuseNet*. This model uses BERT for question feature extraction, which was found to be more effective than XLNet [103]. The authors used two datasets for the development of the model: ImageCLEF 2019 MED-VQA [292] and PathVQA datasets [294].

## 8.5 Image Synthesis

The objective of medical image synthesis is to replace or bypass an imaging procedure that is constrained by time, cost, and labor or to prevent exposure to harmful ionizing radiation from certain imaging modalities. Dalmaz et al. [296] proposed a novel encoder-decoder based generative adversarial network (GAN) model RESVIT for synthesizing missing sequences in multi-contrast MRI and pelvic CT images from source MRI images. The network architecture comprises of a CNN encoder and decoder to leverage the local inductive bias of convolutions and an aggregated residual Transformer as an information bottleneck to learn global representations. RESVIT model synergistically fuses local and global feature representations to achieve superior image synthesis quality. Other GAN-based [297] models such as CycleGAN [298] and CyTran [299] were used to create contrast CT scans from non-contrast CT scans and vice versa. The CyTran architecture incorporates convolutional upsampling, convolution downsampling, and a convolution Transformer block to perform the translation. Kamran et al. [300] proposed VTGAN which combines two generators for looking at local and global features separately with ViT [248] discriminators. It was trained in a semi-supervised manner to synthesize Fluorescein Angiography images [301] along with predicting retinal degeneration. VTGAN successfully synthesized angiogram from fundus images and proved to be robust on spatial and radial transformations.

Yan et al. created MMTrans [302] which uses a Swin-Transformer [252] as both a generator and registration network and a CNN as the discriminator. The generator was used to generate images with the same content as the source modality and the same style as the target modality, while the discriminator was used to measure the similarity between original target modality images and those synthesized by MMTrans. Hu et al. proposed a double-scale graph neural network (GNN) [303] combined with a transformer module to learn long-range dependencies from global features through a discriminator, while for local features, they used CNN. It outperformed established baselines in the IXI dataset. Liu et al. introduced a multi-contrast multi-scale Transformer (MMT) [304] which used missing data imputation as input and proposed a Multi-contrast Shifted Window (M-Swin) to capture intra- and inter-contrast dependencies.

PTNet [305], proposed by Zhang et al., was used to synthesize infant MRI [306] scans. PTNet is a U-net[181] based architecture that incorporates a performer[307] encoder and a decoder with linear space and time complexity. PTNet outperformed previous CNN-based approaches and had a practical execution time of 30 slices per second. Zhang et al. further extended PTNet to 3D MRI as PTNet3D [308] and evaluated it on high-resolution Developing Human Connectome Project (dHCP) [306] and longitudinal Baby Connectome Project (BCP) datasets [309].

## 8.6 Image Reconstruction

Image reconstruction aims to obtain high-quality medical images with minimal cost and risk to the patient.

### 8.6.1 Computed Tomography (CT)

Low-dose computed tomography (LDCT) imaging for clinical diagnosis uses a reduced dose of X-ray radiation compared to conventional CT scans. However, LDCT is prone to noise, which affects the scan quality. Zhang et al. proposed TransCT [310] to enhance the quality of LCDT images using the AAPM-Mayo LDCT dataset [311]. The input image was decomposed into low-frequency and high-frequency components and then content, texture, and high-frequency embeddings were fed to the TransCT model to obtain refined high-frequency textural features. Luthra et al. proposed Eformer [256] which uses a combination of learnable edge-enhancement convolutions called Sobel filters and the LeWin transformer [312] to achieve SOTA performance in denoising LDCT images for detecting metastatic liver lesions (AAPM-Mayo dataset) [311]. Wang et al. [313, 314] proposed convolution-free transformer-based encoder-decoder dilation networks (TED-net) using vanilla transformer blocks for LDCT denoising. Instead of an image, few approaches used informative sinograms generated by restoration modules from origin LDCT images for reconstruction using transformer-based models [315-318].

### 8.6.2 Magnetic Resonance Imaging (MRI)

Korkmaz et al. proposed a MRI reconstruction model based on zero-shot learned adversarial vision Transformer named SLATER [319] to overcome the limitation of data size. Inspired by Deep Image Prior (DIP) [320], they replaced the CNN backbone of DIP with cross-attention transformer structure and finally outperformed DIP both on IXI dataset [269] and multi-coil brain MRI data from fastMRI [321]. Feng et al. [322, 323] introduced a multi-task framework T2Net to share the representations between reconstruction and super-resolution branches. Furthermore, they extended to multi-modalities (MTrans), aiming to learn more knowledge from MRI using both the branches. Fang et al. proposed a cross-modality high-frequency Transformer (Cohf-T) [324] for super-resolving low resolution MR images. Guo et al. proposed a light weight recurrent transformer model ReconFormer [325] which includes pyramid transformer layers [326] to capture intrinsic multiscale information and feature correlation through the recurrent states. Li et al. proposed McMRSR [327] a Transformer based network to model long range dependencies between reference and target images and further aggregate multiscale matched features to reconstruct a target MR image. Few approaches used raw K-space signal of MRI scan instead of final MRI images as they contain learnable information for MRI reconstruction [321, 328-331]. Hu et al. introduced a Transformer-enhanced Residual-error AlterNative Suppression Network [332], which included a regularization term to improve the contribution of high-frequency information during inference. Fabian et al. [333] proposed HUMUS-Net, a two level hybrid CNN Transformer architecture for MRI reconstruction using the fastMRI dataset [321]. Huang et al. [334] proposed a GAN [297] based on Swin-Transformer [252] named ST-GAN, which preserved edge and texture features. Swin-Transformer inspired shifted window attention became the go to Transformer architecture for many studies targeting MRI reconstruction [329, 335-337].

### 8.6.3 Positron Emission Tomography (PET)

PET is a popular imaging technique that measures emissions from radioactively labeled chemicals that were injected into the bloodstream. PET scans can measure metabolic activity and other biochemical functions. Unfortunately, PET suffers from a poor signal-to-noise ratio. Therefore, PET reconstruction requires denoising low-quality PET images to create high-quality ones. Luo et al. proposed a GAN based Transformer model, Transformer-GAN [338] for PET reconstruction with CNN(Encoder)-Transformer-CNN(Decoder) architecture to take advantage of spatial information and long-range dependencies from CNN and transformers respectively. Fu et al. further extended their transGAN-SDAM [339] for fast 2.5D-based L-PET. The transGAN generates higher quality F-PET images followed by the SDAM module which combines spatial information of a F-PET slice sequence to generate whole-brain F-PET images. Jang et al. proposed Spach Transformer v that can leverage spatial and channel-wise information based on local and global MHSA which outperformed baselines on different PET tracer datasets of 18F-FDG, 18F-ACBC, 18F-DCFPyL, and 68GaDOTATATE.

## 9 Transformers for Critical Care

### 9.1 Predicting Long-Term Adverse Outcomes

Yang et al., 2021 predicted a 60-day and 90-day response to targeted immunotherapy of patients with non-small cell lung cancer (NSCLC) using asynchronous clinical time series consisting of chest CT scans, and blood tests, and patient characteristics using an attention module called Simple Temporal Attention [340]. The model predicted which patients would have long-term durable survival gains under an immunotherapy regimen. Similarly, in colorectal cancer, Ho et al., 2021 used Transformer encoders to extract features from sequential measurements of carcinoembryogenic antigen (CEA). It combined CEA measurement features with deep representations of tabular features such as tumor sites, number, dates, and dosage of chemotherapy to predict recurrence [341]. They modified the Transformer to incorporate 1D convolutions prior to localized self-attention [342] Their model outperformed commercial diagnostic tests of colorectal cancer recurrence. Non-clinical population-level claims data has also been modeled using multi-headed self-attention to predict relapse after surgery [343, 344]. These studies utilized the French national health insurance database (SNIIRAM), consisting of health-insurance claims entries of 65 million individuals [345].

## 9.2 Surgical Instruction Generation

Intra-operative surgical assistance AI systems need to solve the task of automatic surgical instruction generation. Zhang et al., 2021 used a transformer-backboned encoder-decoder network combined with self-critical reinforcement learning (RL) to jointly model surgical activity and relationships between visual information and textual description [346]. They used the Database for AI Surgical Instruction dataset (DAISI) to evaluate their model[347]. The authors used a combination of machine translation and image-captioning criteria to evaluate their models, such as BLEU [348], Rouge-L [349], METEOR [350], and CIDEr [351], and SPICE [352]. The combination of Transformer with RL beat baselines comprising LSTM-based fully connected and soft-attention models.

# 10 Transformers for Social Media Data in Public Health

In recent years, using social media data has gained prominence in different areas of public health [353-356]. It is possible to methodically monitor social media posts and Internet information thanks to advances in deep learning and AI. Transformers have been applied to social media data for addressing several public health problems, such as monitoring adverse drug reactions [357, 358], monitoring depression [359], categorizing vaccine confidence [360], and locating disease hotspots [361]. In this section, we present the models used for these purposes and their performance on various datasets.

## 10.1 Monitoring Adverse Drug Reactions (ADRs)

ADR, also known as adverse drug effect (ADE), refers to an undesired, unpleasant and dangerous reaction due to use of a drug [362]. The main steps in monitoring ADRs using social media posts are text classification to find the text that mentions an adverse drug reaction, and the concept and mention extraction of ADE/ADR from the classified text. Breden et al.[357], preprocessed the Twitter dataset from Social Media Mining for Health (SMM4H) 2019 Competition [363] using the lexical normalization [364] method. The best performing model was an ensemble of fine-tuned BERT, BioBERT [55] and ClinicalBERT [60]. In the paper [365] the authors used a more recent dataset provided by SMM4H 2021 [366] for classifying English tweets by concatenating the RoBERTa [96] and ChemBERTa [367] models. The best classification results for Russian tweets were obtained by concatenating the EnRuDR-BERT [368], RuEn training and ChemRoBERTA [362] cross attention. Hussain et al. [369] proposed an end-to-end system based on transfer learning using one prediction head for the text classification, and the other head for labeling the adverse drug responses. The authors fine-tuned BERT with a modular Framework for Adapting Representation Models (FARM), and present the FARM-BERT framework, which gives F-1 score that outperforms competing models on TwiMed-Twitter [370], Twitter [371], PubMed [372], and TwiMed-PubMed [370] datasets. The framework FARM-BERT provides support for multitask learning by combining multiple prediction heads which makes training of the end-to-end systems easier and computationally faster. Raval et al.[358], tackled the same ADE classification problem; however, they framed it as a sequence-to-sequence problem and used the pre-tained T5 model architecture [373] on multiple datasets (SMM4H [374], CADEC [375], ADE corpus v2 [372], WEB-RADT [376] , SMM4H-French [374]). The authors further expanded the proportional mixing and temperature scaling training strategies described in [377] to handle multi-dataset, and present relative improvement on the F-1 score.

## 10.2 Monitoring Depression

Social media provides a vast amount of information for monitoring depression. A large-scale depression dataset on Twitter is presented by [359] and the authors used transformer based models in identifying users suffering from depression using their everyday speech. The importance of psychological test features is also studied when performing depression classification. Some results on the fluctuating depression levels for different groups are also presented.  Matero et al. [378] used pretrained BERT embeddings to encode this information. Kabir et al. [379] presents a dataset observing the severity of depression in tweets, and reported baseline results using BERT and DistilBERT [380].

## 10.3 Monitoring Diabetes

Large-scale Twitter data concerning diabetes related tweets have been collected and used to identify cause-effect relationships [381]. They used a pre-trained BERTweet model [382] to detect causal sentences and a combined BERT+ Random Field Generator model to extract potential cause effect relationships.

## 10.4 Categorizing Vaccine Confidence

Social media plays a key role in engaging people in public relations [383]. Consequently, it provides a great resource to analyze vaccination apprehensions and study the different barriers to the successful vaccinations [384]. One way to do this is by tracking social media conversations about vaccinations. It is essential to be able to annotate the vaccine related content to spot activities that may signal vaccine hesitancy. Kummervold et al. [360], showed that it is possible to get better annotations using state of the art Transformer models compared to the human annotators on maternal vaccination tweets. The performance of neural networks with and without embeddings, LSTMs with GloVe embeddings [52] and without embeddings, default BERT and domain specific BERT are compared with the performance of the human annotators. The domain specific BERT outperformed other methods as well as human annotators.

## 10.5 Locating Disease Hotspot

It is essential to detect disease outbreaks while simultaneously reducing reporting lag time. This can provide an independent source of data to complement traditional surveillance approaches. Alsudias et al. [361] performed a multi-label classification task to identify tweets of infected individuals in the Arabic-speaking world. The authors propose a combination of binary relevance, classifier chains, label power set, multilabel adapted k-nearest neighbors (MLKNN) [385], support vector machine with naive Bayes features (NBSVM) [386], BERT and AraBERT (transformer-based model for Arabic language understanding) [387]. The proposed model achieved an F1 score of up to 88% in the influenza case study and 94% in the COVID-19. It is shown that including informal terms, and non-standard terminology (e.g., the slang term of influenza, symptom, prevention, treatment, infected with) in the encodings improved the performance by as much as 15%, with an average improvement of 8%. The proposed geolocation detection algorithm performed moderately in predicting the location of users according to their tweet content.

# 11 Monitoring Bio-Physical Signals

Transformers have been used to model physical activity, EEG, ECG, and MRI signals from humans. In the following paragraphs we review these works.

## 11.1 Human Activity recognition (HAR)

HAR is a proliferating field of research owing to the recent rise of wearables, smartphones, and Internet of things devices. Some studies have used multi-modal self-attention to fuse features from various modalities in a systematic way [388, 389]. They studied sequences of human movements through multimodal data (such as RGB, depth and skeletal-data) [390-392] or modeled human activity through accelerometer and gyroscope [393-396]. Spatio-temporal bone and joint-sequences from skeleton data have been modeled using multi-scale Transformers using multiple datasets [397-400]. Owing to lack of simple augmentation strategies of longitudinal sensor data, Ramachandra et al., 2021 used Transformer-GAN to provide speedup over existing Recurrent-GAN [401].

## 11.2 Electroencephalograms (EEGs)

EEGs are a widely used non-invasive measurement of brain activity. Transformers have been used to classify visual or motor imagery using EEG signals [402]. It has been shown that extensive self-supervised pre-training using contrastive loss can help Transformer models represent EEG data collected using different hardware, while performing different tasks [403]. Cross-modal Transformers have been used to find contextualized embeddings representing associations between auditory attention detection and EEG signals [404]. This can disentangle sources of brain activity at different time points while the subject is attending to multiple sounds sources simultaneously. Finally, a 2D Transformer was used to capture local self-similarity and feed-forward connections used to capture global self-similarity in a bid to create a novel denoising system for 1D EEG [405].

## 11.3 Electrocardiograms (ECGs)

ECG signals alone and in combination with other sensory information were used to predict stress in subjects using Transformers [406, 407]. Wearable Stress and Affect Detection (WESAD) and SWELL Knowledge Work (SWELL-KW) are publicly available datasets used for this purpose [408, 409]. A transformer network embedded inside a CNN architecture has been used to classify arrythmias [410].

## 12 Transformers for Biomolecular Sequences

Biomolecular sequences can be used to represent genomic, proteomic and drug data. Transformers, being sequence translation models, have been widely used to model the relationships between anomalous biological sequences and corresponding diseases. Moreover, drug/protein synthesis or gene sequence alignment problems have been treated through the lens of machine translation where the Transformer is the model of choice.

### 12.1 DNA

Gene Transformer, which consists of a multi-head self-attention module, automatically detects relevant biomarkers necessary for classifying lung cancer subtypes [411]. It consists of two 1D convolutional layers prior to the MHSA layer to extract low and moderate-level features. A previous study utilized RNA-sequencing values from lung adenocarcinoma (LUAD) and lung squamous cell carcinoma (LUSC) datasets from the Cancer Genome Atlas project [412]. Clauwaert et al. 2020 introduced an attention method that is optimized for nucleotides on top of the Transformer-XL architecture [413]. This attention module included a 1D convolutional layer that extracted overlapping DNA segments of length k called k-mers from the query, key and value matrices of the original DNA sequences. The authors solved three problems including: a) annotating transcription start site (TSS), b) annotating translation initiation site (TIS), and c) recognizing 4mC methylation sites using the following datasets – RegulonDB [414], Ensembl [415], and MethSMRT [416], respectively. A following study utilized comparative TSS annotations from multiple datasets including RegulonDB [414], Etwiller et.al., 2016 (Cappable-seq) [417], Yan et al., 2018 (SMRT-Cappable-seq) [418], and Ju et al., 2019 (SEnd-seq) [419]. In another study, the Transformer-XL network was found to be highly biased towards attending to promoter regions and transcription factor binding sites in the vicinity of the gene under question [420]. Another network, DNABERT was used to predict transcription factor binding (TFB) sites, including proximal and core promoter regions, splice sites, and genetic variants [421]. Reference human genome GRCh38.p13 primary assembly from GENCODE Release 33 [422] was used for pre-training, TATA, and non-TATA promoter data from Eukaryotic Promoter Database (EPDnew) [423] for promoter prediction and ENCODE 690 ChIP-seq profiles from UCSC genome browser [424] were used for predicting TFB sites. Enhancers are regulatory elements that activate promoter transcription over large distances independently of orientation [425]. BERT, pre-trained with masked language modeling (MLM) and next sentence prediction tasks, was combined with 2D convolutions to predict transcription enhancers [426]. The authors used a dataset that describes an enhancer sequencer from nine different cell lines in this study [427, 428].

### 12.2 Protein

Transformers can either predict global properties of protein such as type, function, or cellular localization or infer local properties of selected protein residues such as 2D/3D structure or post-translation modifications (such as phosphorylation and cleavage sites) [429]. The recent success of AlphaFold in practically solving the protein structure prediction problem [430] has proved to be a watershed moment for the application of deep learning to protein problems [431]. However, recent advances in this domain have primarily included fine-tuning pre-trained deep models for learning with small datasets [429].

### 12.3 Molecular Drugs

Transformer have been utilized for the prediction of molecular drugs in many situations as follows.

#### 12.3.1 Drug-Drug Synergy

One of the most useful applications of Transformer networks is in the finding of synergistic combinations of drugs for the treatment of diseases which cannot be cured by a single molecule. The classic example of this is cancer. In cancer, drug combinations alleviate drug resistance and improve therapeutic efficacy. However, the rapidly growing number of anti-cancer drugs makes it extremely resource intensive to search the entire space of synergistic drug combinations. This is where computational models like the Transformer are useful. The TranSynergy model constructed a Transformer model of the cellular effect of drug combinations on different gene-cell line combinations by modeling cell-line gene dependency, gene-gene interaction, and genome-wide drug-target interaction, thereby introducing mechanistic knowledge into the model [432]. The study utilized a large drug synergy score dataset [433] and drug target profiles from DrugBank[434] and ChEMBL[435]. TranSynergy outperformed the SOTA and predicted multiple novel synergistic drug combinations for treating ovarian cancer. Kim et.al., 2020 used multi-task transfer learning to study drug synergy in understudied tissues to overcome data scarcity problems [436]. The authors used a multi-head Transformer network to create an embedding of the Simplified Molecular-Input Line-Entry System (SMILES) representation of drugs. TP-DDI presents a completely end-to-end Transformer pipeline with pretrained BioBERT weights for drug recognition and drug-drug interaction (DDI) classification [437]. This study is conducted on the DDI Extraction 2013 corpus [87] which consists of a list of semantically annotated documents with sentences referring to drugs and DDIs from the DrugBank database and MedLine abstracts.

#### 12.3.2 Drug Synthesis

Transformers have been used to convert the task of target-driven de novo drug-synthesis into a neural machine translation task that converts an amino acid sequence into the chemical formula of its binding drug [438]. This method needs neither any prior information about the drug structure nor the 3D structural information of the protein target. The study used a dataset of binding affinity between proteins and drug-like molecules from the BindingDB database [439]. Synthesized drugs were evaluated on active properties like the number of hydrogen donors/acceptors, molecular weight, length, total polar surface area, number of rotatable bonds, and drug-likeness. Born et.al., 2021 studied the synthesis feasibility of drugs for use against the SARS-Cov-2 virus using a transformer-based retrosynthesis prediction engine [440] consisting of two molecular transformers [441]. They operate on a SMILES representation of a molecule to predict best routes for its synthesis [442]. This information was further utilized by another Transformer model to predict the optimal synthesis protocol using a text representation of the synthesis steps [443]. The approach incorporated variational autoencoders and reinforcement learning to automatically learn molecules that target ACE2, a surface receptor on human epithelial cells that allows entry of the SARS-Cov-2 virus [442].

### 12.3.3 Drug-Target Interactions

In silico drug discovery is driven by computational models of drug-target interactions. Huang et al., developed the Molecular Interaction Transformer which is a transformer-based neural machine which models the interaction space between the most common substructures of molecules and drugs [444]. These substructures were discerned using Frequent Consecutive Sub-sequence algorithm on protein sequences from UniProt dataset[445] and drug SMILES strings from ChEMBL [446]. A Transformer encoder is used to create contextualized embeddings of protein and drug substructures separately which are multiplied to capture their interaction strengths. A CNN extracts higher order interactions from them. Three datasets were employed to learn the transformer and CNN weights- MINER DTI from BIOSNAP [447], BindingDB [448] and DAVIS [449].

Manica et al. [450] proposed an anticancer drug sensitivity model using drug SMILES sequences, gene expression profile of tumors, and protein-protein interaction networks. In this model, an attention-based gene expression encoder generates self-attention weights, a contextual attention layer ingests this gene embedding together with the SMILES encoding of a drug to compute an attention distribution over the SMILES tokens, in the genetic context. CNNs with variable kernel lengths were used to extract information about all possible substructures inside the SMILES sequence. The model outperformed others on a regression task involving prediction of drug IC50 values. Training was done using lenient splitting which prevented cell-drug pairs in the test data from being seen beforehand but did not prevent the model from observing how a given cell interacted with other drugs in the dataset and vice versa. The authors used drug sensitivity data from the publicly available Genomics of Drug Sensitivity in Cancer (GDSC) database for this study [451].

Morris et al. 2020 proposed a transformer-based machine translation method to inform the segmentation of molecular substructures into binding/non-binding a target protein [452]. The authors translated SMILES encodings to IUPAC nomenclatures for a set of 83 million compounds from PubChem [453] database and used the resultant cross-representation attention embeddings as features to classify binding/non-binding compartments of molecules from BindingDB [439] to important proteins including HIV-1 protease.

### 12.3.4 Drug Metabolism Prediction

Metabolic processes in the human body can change a drug's structure, therefore diminishing its safety and efficacy. Therefore, investigation of the metabolic fate of a candidate drug is an essential component of drug design studies. Litsa et al., 2020 fine-tuned a pretrained Molecular Transformer, and used an ensemble of them with beam search to find k-likeliest metabolites from every drug [454]. The Molecular Transformer [441] was pretrained on this dataset [455] consisting of 900,000 training instances. The network was further fine-tuned using a manually curated dataset combining samples from Drug-Bank (version 5.1.5) [434], Human Metabolome Database (HMDB) (version 4.0) [456], HumanCyc from MetaCyc (version 23.0) [457], Recon3D (version 3.01) [458], the biotransformation database (MetXBioDB) [459] and reaction rules from SyGMa [460]. Their network outperformed SOTA models including the BioTransformer [459].

## 13 Discussion

This paper presented an exhaustive summary of Transformer-based applications in healthcare for tasks such as clinical report generation, medical image segmentation and registration, molecular sequencing, drug-drug interactions, protein synthesis, surgical augmentation, and bio-physical signal analysis. Although relatively new, Transformers have been rapidly adopted owing to their inherent ability to capture long-range dependencies in the data. This is bolstered by the fact that most bio-medical entities can be represented by interaction networks, which are characterized by long-range dependencies. However, the parallelizable attention module at the heart of the Transformer network is computationally expensive and often needs to be optimized for efficient usage. In what follows, we highlight potential drawbacks of transformers, how to overcome them, and new directions enabled by Transformers.

### 13.1 Interpretability and Explainability

Most deep learning systems are considered "black box" models because their inferences do not come with any discernable explanation. This lack of interpretability has traditionally prevented the systemic acceptance of AI-aided diagnostics in the medical domain. Transformers inherently provide some transparency through visualization of their attention weights. Trained attention weights elucidate contextual information significant for downstream inference. However, Chefer et al. [461] show that Transformer attention is often fragmented and does not provide a robust explanation. Interpreting Transformers is also challenging due to the frequent use of skip-connections and the dynamic nature of the model, which involves weight computation through matrix multiplication. Therefore, Transformer interpretability, albeit being an inherent property, is not trivial. In case of vision Transformers, Bohle et al. [462] proposed B-cos transfomers, for holistic exaplainations for their decisions while retaining the performance to the baseline ViTs. Disease diagnosis prediction studies [463, 464] have generated attention visualizations and cosine similarity between the learnt clinical diagnoses embeddings verified by expert clinicians to understand whether the trained model could capture the underlying semantic of diagnoses codes. However, there remains a need to develop novel techniques to improve the interpretability of Transformer models tailored towards healthcare AI.

### 13.2 Environmental Impact

Advances in AI in recent years have come at the cost of a massive carbon footprint. Training a large-scale deep learning model is estimated to produce 626,000 lbs of carbon dioxide, equivalent to five automobiles' lifetime emissions [465]. The number of computational resources researchers use to create SOTA models has doubled every three to four months [466]. Most emissions are associated with developing and training deep learning algorithms, whereas finetuning and adaptation contribute less [467]. Strubell et al. [465] suggested that researchers report hardware-independent training time measurements, such as the number of gigaflops required for training convergence and measuring model sensitivity to data and hyperparameters. The last decade has seen advancements in AI-augmented healthcare, on the one hand, and carbon emissions caused by AI systems that are detrimental to the climate and public health on the other. Large healthcare conglomerates and governmental agencies around the world should target net-zero carbon emissions. United Kingdom National Health Service has set a goal of net-zero emissions by 2040 [468]. Goals such as this are vital to promote the development of energy-efficient hardware and algorithms that make AI sustainable and globally accessible.

### 13.3 Computational Costs

The reason behind the impact of Transformers is their high parametric complexity, flexibility to handle unequal input lengths and model scalability. However, Transformers' ability to be trained on enormous datasets comes with expensive computational training budgets. The LLM GPT3 [25] by OpenAI training is estimated to cost $4.6 million and 355 years of computing time using the Nvidia Tesla V100 device [469]. Google's 530 billion parameters PaLM model is estimated to consume 103,500 KWh over 60 days [470]. Training and deploying large-scale AI models with high-end hardware requirements in healthcare settings is challenging. For example, for on-premise use in a hospital, a centralized compute cluster similar to ChatGPT might need to be maintained and interacted with using an API. However, healthcare settings typically need lightweight models to generate real-time predictions with minimal maintenance costs. Techniques for compressing deep learning models, such as pruning [471], knowledge distillation [472], and quantization [473], can be used to provide a more efficient model implementation for deployment within practical hardware constraints.

#### 13.3.1 Model Compression

Transformer models can be efficiently compressed by discarding some attention heads during the inference phase. Michel et al. [474] showed that models trained on multiple heads during training time need not require all the heads during test time. Similar redundancy has been observed in generating attention matrices from multiple heads [475].

#### 13.3.2 Quantization

Quantization-based approaches reduce the number of bits/unique values required to represent model weights and intermediate layer activations. There has been growing interest among researchers in recent years in quantizing transformer networks. Shen et al. [476] observed ~2.3% degradation in performance with quantization down to 2 bits, corresponding to 13X compression of network parameters and 4X compression on embeddings and activations. It was observed that position embedding and the embedding layers are more sensitive to quantization than other operations.

#### 13.3.3 Knowledge Distillation

The knowledge distillation approach aims to train small networks (aka student) using the knowledge from the large models (teacher). Student models are obtained by reducing encoder width, number of heads, and number of encoders and replacing them with CNN, BiLSTM, or a combination [477]. Dimensional incompatibility between the student and teacher due to compact representations can be overcome by projecting teacher or student outputs [478].

### 13.4 Fairness and Bias

A model is biased when it exhibits undesired dependence on an attribute of the data that belongs to a specific demographic group [479], and could lead to unfavorable treatment of particular patient groups. Researchers have observed that bias often arises when the datasets used to train the models under-represent certain patient populations [480, 481]. Although this is a prevalent bias problem during training, other sources of bias at all stages exist, including during problem formulation, data collection, data preprocessing, model development and validation, and model deployment (e.g., due to unmonitored drift) [482]. With the increasing scale of models and amount of data available, the existing biases and stereotypes perpetuate into the models leading to unfair and biased outcomes [50]. Thorough validation should be done before deploying the model to evaluate the performance of underrepresented groups. The models should be continuously monitored and audited for fairness and bias post-deployment.

### 13.5 AI Alignment

The goal of AI alignment is even broader than preventing bias by striving to design AI systems that align with human values and goals. An AI system is considered aligned when the system behaves in ways beneficial to humans while minimizing the risk of unintended consequences and harmful outcomes. LLMs sometimes confidently assert false claims that do not reflect facts, a phenomenon termed hallucination [483]. These hallucinations by the nonaligned models fail to meet the user's expectations of correct answers faithful to the existing sources. Ensuring AI systems are aligned with human values and goals is challenging because predicting and designing for every potential desired and undesired outcome can be hard. As AI systems become more capable, they become increasingly susceptible to the alignment problem, which can result in unintended and harmful consequences [484]. AI alignment is especially critical in healthcare when deploying large-scale foundation models to ensure these models are ethical, responsible, respectful of patient privacy, and, most importantly, not causing harm. Healthcare professionals and the AI research community need to develop a clear set of standards and guidelines to establish ethical use of AI in health care.

### 13.6 Data Privacy and Data Sharing

Preserving patient privacy is a required feature in all healthcare AI systems. Federal regulations based on the Health Insurance Portability and Accountability Act (HIPAA) regulate the development of AI models that use patient information [485, 486]. Nonetheless, this also adversely impacts the development of large models such as Transformers that require large amounts of data. Utilizing data from a few sources, such as select public repositories, can skew the model inferences based on underlying limitations in dataset collection (different equipment, protocol, and cohort demographics), processing (specific heuristic or statistical preprocessing), and deployment (different metadata, availability, and maintenance). These biases can skew predictions that favor or adversely affect certain population groups over others, leading to a degradation in the quality and equity of healthcare for individuals from the protected group and stymieing the research on age, sex, or race-related medical conditions.

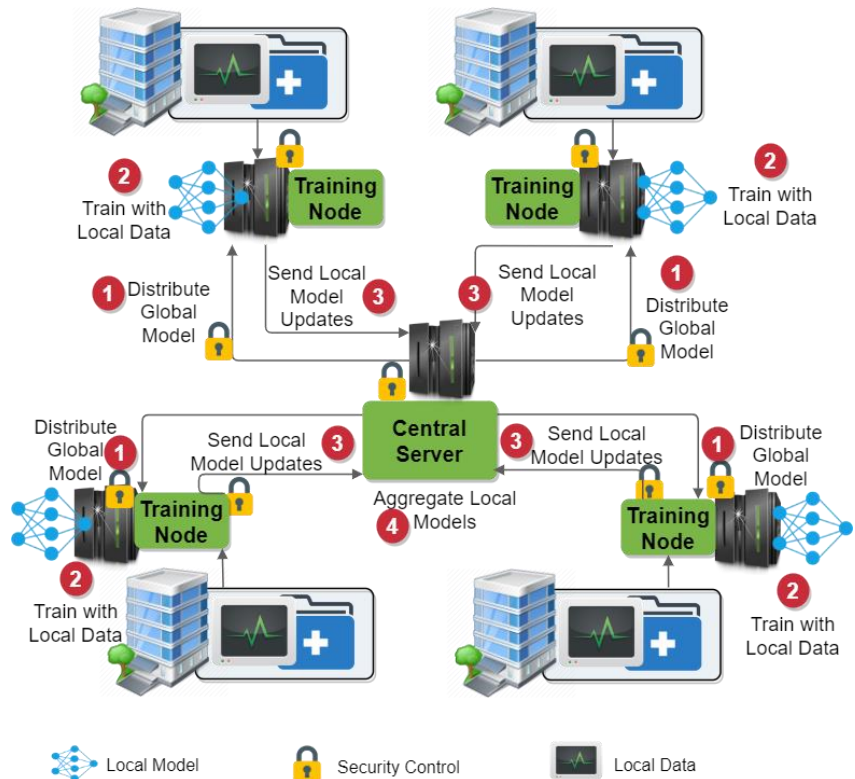

Figure 11. Schematic of Federated learning with a central server that interacts with training nodes at different locations continuously updating the model parameters without exchanging the data between local and central servers.

The Federated learning (FL) paradigm shown in Fig 11 aims at developing a shared training model that can leverage data from multiple fragmented sources, such as different healthcare institutions, without divulging sensitive patient information [487]. FL communicates between various data sources by exchanging model-specific characteristics like parameters and gradients without exchanging patient information directly. Recent efforts in FL have targeted digital health objectives like determining patient clinical similarity [488, 489], mortality and ICU length-of-stay [490], brain segmentation [491], and brain-tumor segmentation [492, 493]. FL can perpetuate many healthcare innovations in the future. However, there are technical challenges in building an operational FL workflow, such as inhomogeneous data distributions, computational hardware differences, inconsistent privacy preservation settings, and resultant performance trade-offs [494].

## 14 Conclusion

Transformer models have demonstrated enormous potential in a wide variety of healthcare applications. They possess a unique ability to model various data modalities, including images, clinical text, biophysical signals, and genomic data. From disease diagnosis to drug discovery, Transformer models exhibit the potential to improve patient outcomes and advance medical research. However, various challenges and limitations remain to be addressed before they are widely accepted into regular clinical practice. These include data limitations, biases, privacy, security, and truthfulness. The majority of the models currently in use are task-specific, and there is a need to utilize robust multimodal inputs in many cases. Nevertheless, the future of AI in healthcare is optimistic, with promising advancements and opportunities presented by large-scale transformer models.